\documentclass[10pt,letterpaper]{article}
\usepackage{algorithm2e}
\usepackage{amsfonts}
\usepackage{amsmath}
\usepackage{amssymb}
\usepackage{bbm}
\usepackage{bm}
\usepackage{chemfig}
\usepackage{chemmacros}
\usepackage{color}
\usepackage{enumitem}
\usepackage{fullpage}
\usepackage{geometry}
\usepackage{graphicx}
\usepackage[hidelinks]{hyperref}
\usepackage[utf8]{inputenc}
\usepackage[version=4]{mhchem}
\usepackage{physics}
\usepackage{subcaption}
\usepackage{systeme}
\usepackage{svg}
\usepackage{textgreek}
\usepackage{tikz}
\usepackage{verbatim}

\usetikzlibrary{arrows.meta, positioning, shapes.geometric}
\tikzstyle{arrow} = [very thick,->,>=stealth]
\tikzstyle{compartment} = [circle,
						   minimum width=1cm,
						   text centered,
						   draw=black,
						   fill=white,
						   very thick]
\tikzstyle{source} = [rectangle,
					  minimum width=1cm,
					  text centered,
					  draw=black,
					  fill=white,
					  very thick]

\newcommand{\ie}{\emph{i.e.}}
\newcommand{\eg}{\emph{e.g.}}
\newcommand{\etal}{\emph{et al.}}

\let\eps\varepsilon
\newcommand{\vect}[1]{\boldsymbol{#1}}
\newcommand{\NN}{\text{NN}}
\newcommand{\R}{\mathbb{R}}
\renewcommand{\norm}[1]{\left\lVert#1\right\rVert}

\newcommand{\biot}{\text{B}}

\newcommand{\E}{\ensuremath{\ce{E}}}
\newcommand{\ESzero}{\ensuremath{\ce{ES0}}}
\newcommand{\ESone}{\ensuremath{\ce{ES1}}}
\newcommand{\Szero}{\ensuremath{\ce{S0}}}
\newcommand{\Sone}{\ensuremath{\ce{S1}}}
\newcommand{\Stwo}{\ensuremath{\ce{S2}}}
\renewcommand{\S}{\ensuremath{\ce{S}}}
\newcommand{\Stot}{\ensuremath{S_{\text{tot}}}}
\newcommand{\Etot}{\ensuremath{E_{\text{tot}}}}

\newcommand{\cE}{\ensuremath{[\ce{E}}]}

\newcommand{\cESzero}{\ensuremath{[\ce{ES0}]}}
\newcommand{\cESone}{\ensuremath{[\ce{ES1}]}}

\newcommand{\cSzero}{\ensuremath{[\ce{S0}]}}
\newcommand{\cSone}{\ensuremath{[\ce{S1}]}}
\newcommand{\cStwo}{\ensuremath{[\ce{S2}]}}

\newcommand{\kf}[1]{\ensuremath{k_{\text{f},#1}}}
\newcommand{\kr}[1]{\ensuremath{k_{\text{r},#1}}}
\newcommand{\kcat}[1]{\ensuremath{k_{\text{cat},#1}}}

\newcommand{\kfx}{\ensuremath{k_{\text{f}}}}
\newcommand{\krx}{\ensuremath{k_{\text{r}}}}
\newcommand{\kcatx}{\ensuremath{k_{\text{cat}}}}
\newcommand{\keffx}{\ensuremath{k_{\text{eff}}}}

\title{On the Parameter Combinations that Matter\\and on Those that do Not}
\author{Nikolaos Evangelou$^{1,\dagger}$, Noah J. Wichrowski$^{2,\dagger}$, George A. Kevrekidis$^3$,\\Felix Dietrich$^4$, Mahdi Kooshkbaghi$^5$, Sarah McFann$^{6,7}$, Ioannis G. Kevrekidis$^{1,2,\star}$}
\date{}

\begin{document}
	\maketitle
	
	\begin{abstract}
		\noindent We present a data-driven approach to characterizing nonidentifiability of a model's parameters and illustrate it through dynamic as well as steady kinetic models. By employing Diffusion Maps and their extensions, we discover the minimal combinations of parameters required to characterize the output behavior of a chemical system: a set of \emph{effective parameters} for the model. Furthermore, we introduce and use a Conformal Autoencoder Neural Network technique, as well as a kernel-based Jointly Smooth Function technique, to disentangle the \emph{redundant} parameter combinations that do not affect the output behavior from the ones that do. We discuss the interpretability of our data-driven effective parameters, and demonstrate the utility of the approach both for behavior prediction and parameter estimation. In the latter task, it becomes important to describe level sets in parameter space that are consistent with a particular output behavior. We validate our approach on a model of multisite phosphorylation, where a reduced set of effective parameters (nonlinear combinations of the physical ones) has previously been established analytically.
	\end{abstract}
	\vspace{10pt}
	
	{
	\noindent\textsuperscript{1} Department of Chemical and Biomolecular Engineering, Johns Hopkins University
	
	\noindent\textsuperscript{2} Department of Applied Mathematics and Statistics, Johns Hopkins University
	
	\noindent\textsuperscript{3} Department of Mathematics and Statistics, University of Massachusetts, Amherst
	
	\noindent\textsuperscript{4} Department of Informatics, Technical University of Munich
	
	\noindent\textsuperscript{5} The Program in Applied and Computational Mathematics, Princeton University
	
	\noindent\textsuperscript{6} Department of Chemical and Biological Engineering, Princeton University
	
	\noindent\textsuperscript{7} Lewis-Sigler Institute for Integrative Genomics, Princeton University\\
	
	\noindent$^\dagger$ N.E. and N.J.W. contributed equally to this work.

	\noindent$^\star$ Corresponding Author: \href{mailto:yannisk@jhu.edu}{yannisk@jhu.edu}
	}
	
	\vspace{15pt}\noindent\textbf{Key words:} manifold learning, parameter nonidentifiability, model order reduction, data mining
	
	\pagebreak
	\section{Introduction}\label{sec:introduction}
	Model reduction has long been an important endeavor in mathematical modeling of physical phenomena and, in particular, in the modeling of large, complex kinetic networks of the forms that arise in combustion or in cellular signaling~\cite{goussis2011model,kopf2021latent,snowden2017methods}. A rich array of techniques, often based on time-scale separation, exist that can result in a smaller number of \emph{effective state variables} and, consequently, a reduced set of coupled nonlinear differential equations (\eg, Benner \etal~\cite{benner2015survey}, Quarteroni \etal~\cite{quarteroni2014reduced}, and from our work~\cite{chiavazzo2014reduced,deane1991low,foias1988computation,nadler2006diffusion,shvartsman2000order}). Yet it also becomes important to discover, when possible, a smaller number of \emph{effective parameters}. These are (possibly nonlinear) combinations of the original, usually physically meaningful, model parameters on which the output behavior depends. A universally accepted and practiced approach towards reducing the set of parameters, undertaken before any computation is, of course, dimensional analysis~\cite{barenblatt1996scaling}. 
	
	Beyond dimensional analysis, the issue of parameter nonidentifiability, whether truly structural or approximate, has been the subject of extensive studies for decades, with rekindled interest in recent years~\cite{cole2020parameter,raue2009structural}. Such developments are eloquently summarized in~\cite{brouwer2018underlying}. This can be attributed in part to sloppiness/MBAM studies~\cite{gutenkunst2007universally,transtrum2014model}; the study of active subspaces~\cite{constantine2015active}; the increased availability and exploitation of symbolic regression packages~\cite{riolo2011genetic}; and, more generally, to recent advances in data science and manifold learning techniques~\cite{coifman2006diffusion,holiday2019manifold}. To a large extent, established model reduction techniques hinge on the availability of analytical model equations and operations (\eg, singular perturbation theory based expansions) on these closed form equations.
	
	This work aspires to synthesize and implement a purely data-driven process for finding reduced effective parameters. The type of models we consider here are systems of coupled, nonlinear, first-order differential equations describing time-evolution of chemical/biochemical reaction networks,  but the approach is applicable more generally to the parameterization of input-output relations. Here, the inputs are the parameters, and the outputs are time series of the system observables, such as species concentrations, temperatures, or functions of such quantities.
	
\begin{figure}[ht]
		\centering
		\includegraphics[width=0.8\textwidth]{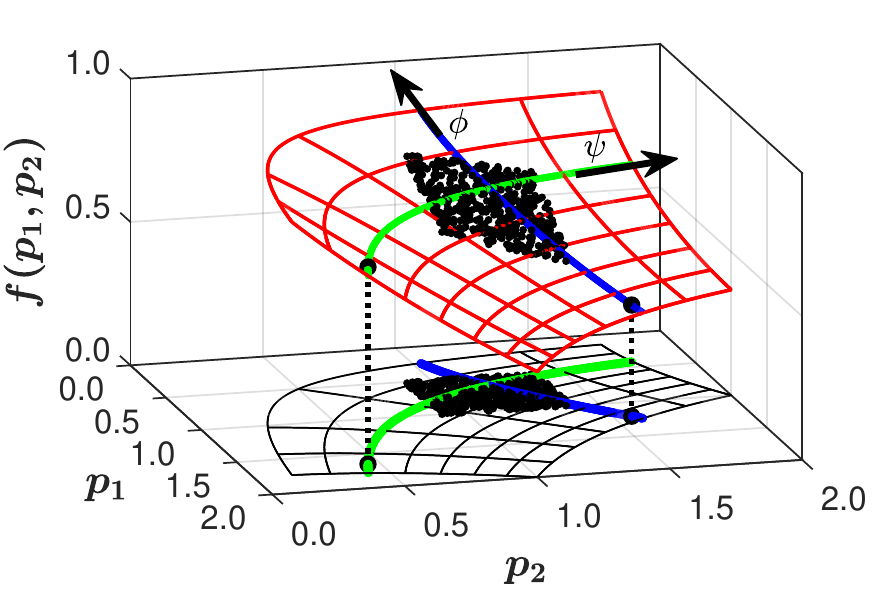}
		\caption{The function $f(\phi)=\exp(-\phi/2)$, with $\phi=p_1p_2$, is sampled at a cloud of points and plotted against the two parameters $p_1,p_2$ (red mesh). Here, $\phi$ is the \textit{effective parameter}, which we call the ``meaningful'' parameter combination. The green curve indicates a level set of this effective parameter, for which ${f(\phi)-C=0}$ for some constant $C$ (here $C=0.75$). The blue curve illustrates the direction orthogonal to each level set of $\phi$, parameterized by $\psi=p_1^2-p_2^2$, which we call the ``redundant'' parameter combination because it does not affect the output. The projection onto the $(p_1,p_2)$-plane helps illustrate the level sets of the meaningful and redundant parameter combination(s) in parameter space.}
		\label{fig:2D-illustration}
	\end{figure}

	In Figure~\ref{fig:2D-illustration}, we illustrate a simple model with structurally nonidentifiable parameters. The model output, $f(p_1,p_2)=\exp(-p_1p_2/2)$, though plotted as function of the two parameters $(p_1,p_2)$, in fact depends only on their product $\phi=p_1p_2$. The output data do not suffice to identify or estimate $p_1$ and $p_2$ independently: observations can only confine pairs of $p_1$ and $p_2$ to a level set, colored green in Figure~\ref{fig:2D-illustration}, of the \emph{effective parameter} $\phi$. It is interesting to observe that these level sets are parameterized by the quantity $\psi=p_1^2-p_2^2$, which is conformal everywhere to $p_1p_2$, thus making $\phi$ and $\psi$ an orthogonal system of coordinates (cf. polar or hyperbolic coordinates). A level set for $\psi$ is colored blue in Figure~\ref{fig:2D-illustration}. This is the parameter combination that \emph{does not matter to the output}, one that is ``redundant:'' keeping the output constant while changing $\psi$ traces out the level set $\phi=C$. To trace out the possible values of the output, one could of course fix one parameter (say, $p_2$) and vary the other(s). In that case, however, the sensitivity of the output to the variation of $p_1$ depends on the value at which we choose to keep $p_2$ constant. This variability is avoided when using a conformal orthogonal set of coordinates, such as the one in the figure. 
	
	In our illustrative models, the system is available in the form of a ``black box'' set of ODEs: given parameter values and initial conditions, one can record time series of the evolution of the system states, or of functions of the system states. But the evolution equations are not explicitly available, so that analytical (possibly perturbative) approaches to reduction of either system states or parameters (outputs or inputs) cannot be undertaken. Given such an input-output model, we start by systematically prescribing a set of numerical experiments for data collection. These data will be processed with manifold learning techniques---here, Diffusion Maps (DMaps) and Geometric Harmonics (GH)---as well as their extensions: output-informed DMaps and Double DMaps GH. Processing the data will:
	\begin{itemize}
		\item Determine the number of model parameter combinations that matter, \ie, the meaningful effective parameters that affect the model output;
		\item Consequently, determine the number of model parameter combinations that do not matter, the redundant ones; 
		\item Interpret the meaningful parameter combinations through computational testing/validation of expert suggestions, or possibly through symbolic regression;
		\item Disentangle the redundant parameter combinations from the meaningful effective ones~\cite{achille2018emergence,locatello2019challenging}, which is accomplished using deep learning techniques (Conformal Autoencoders) or, alternatively, kernel-based Jointly Smooth Feature extraction~\cite{dietrich2020spectral}; and
		\item Translate between the data-driven effective parameters and physical ones, which underscores the importance of level sets in parameter space consistent with the same output behavior.
	\end{itemize}
	We believe these capabilities constitute a useful toolkit for data-driven reparameterization of models, whether computational or physical/experimental. In the experimental setting the same toolkit can be applied; one will perturb (``jiggle'') all inputs/parameters around a base point, record the richness of the resulting output behavior, and establish (through the same framework) correlations between parameters' richness and output  richness---quantify it and parametrize it.
	
	The remainder of the paper is organized as follows: In Section~\ref{sec:parameter-reduction}, we will demonstrate and visualize the discovery of the intrinsic dimensionality of the meaningful effective parameter space through our main illustrative example: a six-equation multisite phosphorylation (MSP) kinetic model and its analytical reduction by Yeung \etal~\cite{yeung2020inference}. In Section~\ref{sec:parameter_identification}, we compare our data-driven effective parameter constructs with those previously obtained analytically and discuss their interpretability, both numerically and through symbolic regression. Finally, we demonstrate the use of these effective parameters in behavior prediction for new physical parameter settings in Section~\ref{sec:behavior_prediction} and (a type of) parameter estimation for previously unobserved behaviors in Section~\ref{sec:parameter_estimation}. Towards the latter task, in Section~\ref{sec:dont_matter_level_sets}, we discover and parameterize entire level sets in parameter space that are consistent with this new observed behavior; this requires discovering the redundant parameter combinations. In Section~\ref{sec:JSF}, a deep learning architecture (Conformal Autoencoder Networks) as well as an alternative kernel-based \emph{Jointly Smooth Functions extraction} are used for this task of disentangling meaningful effective parameters from redundant ones. We conclude by summarizing the approach and offering a discussion of its potential, shortcomings, and current research directions.
	
	In Appendices~\ref{sec:compartmental_model} and~\ref{sec:eta_phi_biot}, we have included two additional examples. The first comes from a textbook nonidentifiable dynamical system representing a compartmental model; the second is a steady state example, which allows us to illustrate how our data-driven framework behaves when transitions between qualitatively distinct behavior regimes arise as one traverses the original parameter space. 
	
	\section{Results}\label{sec:results}
	The ``black box'' models that we seek to parameterize in our data-driven work arise mainly from chemical kinetic mechanisms, (\eg, Equation~\ref{eqn:MSP_mechanism}), which give rise to systems of ODEs for the evolution in time of the species concentrations as output, depending on several kinetic parameters, possibly including the total quantity of a catalyst or enzyme, as input. In certain parameter regimes, the existence of disparate (fast-slow) time scales allows one to explicitly reduce a detailed kinetic scheme through, \eg, the Bodenstein~\cite{bodenstein1913theorie} or Quasi-Steady-State Approximation (QSSA) to an effective reduced one, characterized by new, reduced effective parameters.
	
	The detection of such effective parameters in our scheme will be achieved by using the manifold learning algorithm DMaps~\cite{coifman2006diffusion}, for which a more detailed description is given in Appendix~\ref{sec:diffusion_maps}. We will illustrate that, given a systematically collected data set, and with an appropriate \textit{metric}, DMaps can be used for parameter reduction: discovery of effective parameter combinations that affect the output, as well as parameter combinations that do not affect it. The motivation for our work arose from studying the reduction of the following Multisite Phosphorylation (MSP) Model.
	
	\subsection{The Multisite Phosphorylation Model}\label{sec:msp_model}
	Yeung \etal~\cite{yeung2020inference} proposed and analyzed a kinetic model that describes the dual phosphorylation of Extracellular Signal Regulated Kinase (ERK) by an enzyme known as MEK. Here, ERK can exist in any of three states: $\Szero$, $\Sone$, and $\Stwo$, where the subscript indicates the number of times the substrate has been phosphorylated. The MEK enzyme, denoted by $\E$, forms complexes $\ESzero$ and $\ESone$ with the first two phosphostates. The reaction mechanism for this system is given by
	\begin{equation}
		\label{eqn:MSP_mechanism}
		\schemestart
		\chemfig{E} \+ \chemfig{S_0}\arrow(aa--bb){<=>[\kf1][\kr1]}\chemfig{ES_0}
		\arrow(bb--cc){->[\kcat1][]}\chemfig{ES_1}
		\arrow{->[\kcat2][]} \chemfig{E} \+  \chemfig{S_2}
		\arrow(@cc--xx1){<=>[*0\kr2][*0\kf2]}[-90]\chemfig{E} \+ \chemfig{S_1}\;, 
		\schemestop
	\end{equation}
	with the six rate constants comprising our vector of inputs/parameters: $$\vect{p}=[\kf1,\,\kr1,\,\kcat1,\,\kf2,\,\kr2,\,\kcat2]^\top\in\R^6\,.$$ The governing system of ordinary differential equations is provided in Appendix~\ref{sec:yeung_model}.
	
	Yeung \etal~used the QSSA for the species $\ESzero$ and $\ESone$ along with stoichiometric conservation to approximately reduce the above system: if the assumptions
	\begin{equation*}
		\Stot\ll\frac{\kf1+\kcat1}{\kf1}\,,\qquad \Stot\ll\frac{\kr2+\kcat2}{\kf2}
	\end{equation*}
	reasonably hold, where $$\Stot=\left.\cSzero\right|_{t=0}=\cSzero+\cSone+\cStwo+\cESzero+\cESone\,,$$ then the initial model reduces to a three-state linear kinetic model that depends on only three effective parameters, which are combinations of the full model parameters:
	\begin{align}
		\label{eqn:stas_effective_parameters}
		\kappa_1 & =\cE\,\frac{\kf1\kcat1}{\kr1+\kcat1}\,,\nonumber\\
		\kappa_2 & =\cE\,\frac{\kf2\kcat2}{\kr2+\kcat2}\,,\\
		\pi & =\frac{\kcat2}{\kr2+\kcat2}\,.\nonumber
	\end{align}
	The reduced equations can be found in Appendix~\ref{sec:reduced-msp}. We will attempt to derive such a reduced parameterization in a data-driven manner.
	
	\subsection{Data-Driven Parameter Reduction}\label{sec:parameter-reduction}
	We select a \emph{base point} in parameter space,
	\begin{equation*}
		\tilde{\vect{p}}=[0.71,\,19,\,6700,\,9200,\,0.97,\,5200]^\top\,,
	\end{equation*}
	which is situated in the region of parameter space where the reduction assumptions hold. We select a reference initial condition $\cSzero=5$ and $\cE=0.66$, with the other species not initially present. Numerically integrating the associated system of ODEs, we collected $10,000$ dynamic observations of the system output in response to perturbations of each parameter within $\pm10\%$ of its base value.Note that our random parameter perturbations, are merely a device for sampling the neighborhood of the base point in input/parameter space; a grid of equally-spaced points would also suffice.
	
	In the following analysis, we take as our outputs the concentration $\cStwo$ at $t\in\{2,4,\ldots,20\}$, which yields a 10-dimensional observation vector at each parameter setting. For this example, the choice of $\cStwo$ as the observed output is not particularly significant; the temporal response of \emph{any} time-varying chemical species or combination thereof would give the same results (based on Takens' embedding theorem~\cite{takens1981detecting}). We will refer to this particular data set $\textbf{X}$ as the \textit{transient data}. This data set samples what in the literature is referred to as \emph{the model manifold}, whose dimensionality determines the number of meaningful (effective) parameters~\cite{sethna2013sloppiness,transtrum2010nonlinear,transtrum2014model}. 
	
	A second data set, $\textbf{Y}$, was obtained through computational optimization experiments, in which we estimated vectors of six parameter values that best fit the reference transient we obtained at the base parameter setting. In these experiments, initial conditions were chosen randomly from a log10-uniform distribution, with lower and upper bounds set, respectively, at $10^{-3}$ and $10^{+3}$ times the rates estimated by Aoki \etal~\cite{aoki2013quantitative}. We performed nonlinear least-square fits of these transients from 1,000 random initial conditions in 6D parameter space, as described in~\cite{yeung2020inference}. Upon successful completion of these computations, we have 1,000 six-dimensional ``optimal fits'' of the base parameter setting; we call this data set the \textit{optimization data}.
	
	We first computed \textit{output informed} DMaps, with the distance metric described in Appendix~\ref{sec:diffusion_maps}, on the transient data set. The number of independent/non-harmonic eigenvectors indicates the effective dimensionality of the model manifold. We found three non-harmonic DMaps eigenvectors~\cite{dsilva2018parsimonious}, $$\vect{\phi}=(\phi_1,\phi_3,\phi_9)\in\Phi\subset\R^3$$ and deduced that the intrinsic dimensionality of the transient data set, and thus of the model manifold, is three. We then turned to the optimization data set and performed both Principal Component Analysis and ``regular'' DMaps. We found that the intrinsic dimensionality of the optimization data set is also three, whether we estimate it from PCA or from DMaps. These two results corroborate/complement each other, since three plus three equals six, the total number of original parameters. 
	
\begin{figure*}[htbp]
		\centering
		\includegraphics[width=0.9\textwidth]{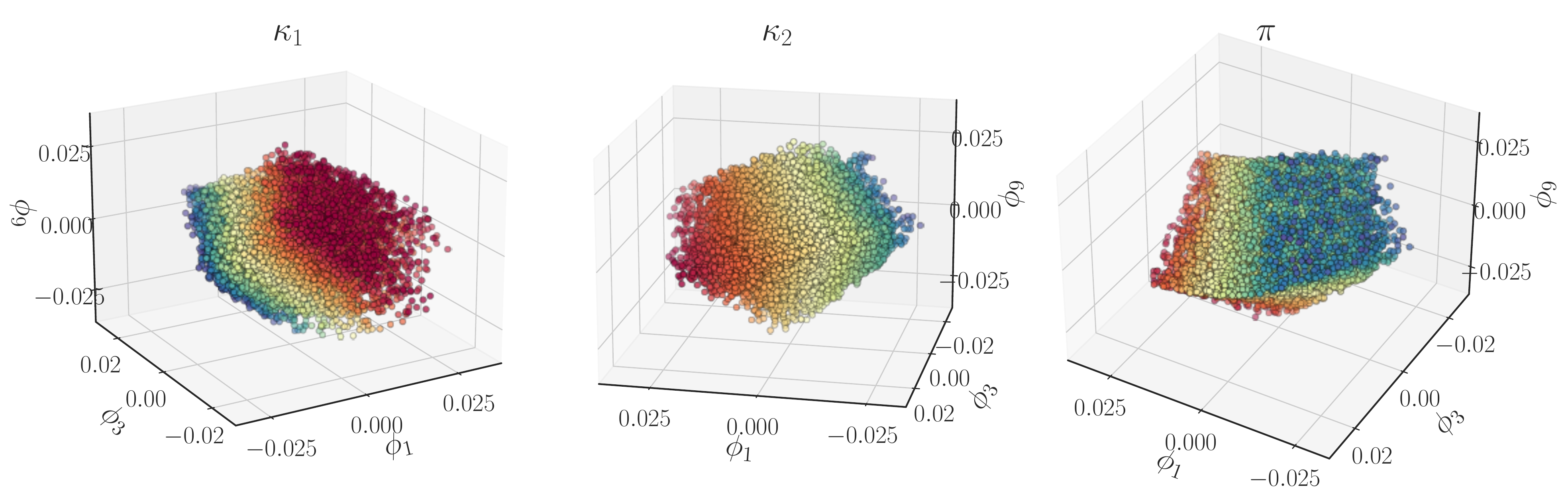}
		\caption{The first three independent, nontrivial eigenvectors, $\vect{\phi}_1,\vect{\phi}_3,\vect{\phi}_9$, colored by (computed) values of the three theoretical effective parameters, $\kappa_1,\kappa_2,\pi$, respectively, for a transient data set.}
		\label{fig:DMaps_coordinates_parameters}
	\end{figure*}

	The dimensionality of the \emph{transient data set} could be estimated from the dimension of the null space of either the sensitivity matrix or the sensitivity Fisher information matrix~\cite{brouwer2018underlying} at the base point. Beyond this estimate, however, our approach discovers a \emph{global} parameterization over the data of the output in terms of $\vect{\phi}=(\phi_1,\phi_3,\phi_9)$, which are our data-driven effective parameters. These eigenmodes capture the directions, in full parameter space, that matter to the output: the parameter changes that affect the response of our system. Figure~\ref{fig:DMaps_coordinates_parameters} illustrates these three leading non-harmonic eigenvectors, colored by the analytical effective parameters of Yeung \etal~in Equation~\ref{eqn:stas_effective_parameters}. Even though it is difficult to visually appreciate a 3D point cloud through color, we believe one gets a clear visual impression that the data driven effective parameter set and the analytical effective parameter set are one-to-one with each other. We will quantify this below.
	
	We remind the reader that the DMaps effective parameters, like the analytical ones, will in general correspond to \emph{combinations} of the original parameters. But while the analytical effective parameters are physically explainable (Equation~\ref{eqn:stas_effective_parameters} shows their dependence on the original parameters), no such \emph{a priori} physical interpretation comes with the proposed data driven effective parameters. We will address this issue below.
	
\begin{figure}[ht]
		\centering
		\includegraphics[width=0.8\textwidth]{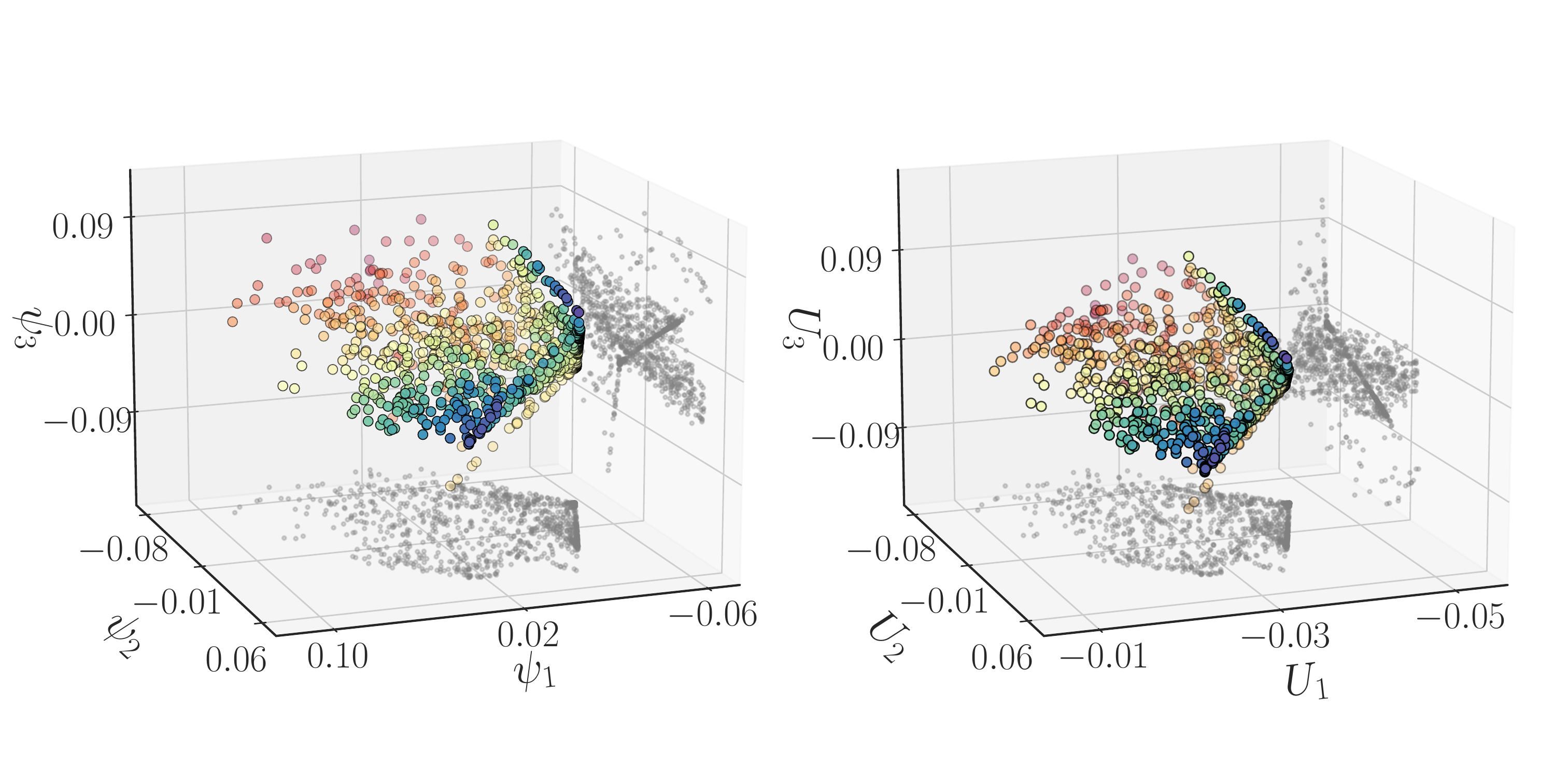}
		\caption{[Left] independent eigenvector coordinates, ${\psi}_1,{\psi}_2,{\psi}_3$, for the optimized data set, colored by $\psi_2$. [Right] the three dominant singular vectors computed with PCA, colored by the second, $U_2$.}
		\label{fig:PCA_vs_DMAPs}
	\end{figure}

	Computing DMaps on the \emph{optimization} data also results in an intrinsically three-dimensional parameterization of the manifold of equivalent optima, Figure~\ref{fig:PCA_vs_DMAPs}. The intrinsic parameters computed for this data set uncover the directions in parameter space that produce (approximately) the same response: the reference trajectory at the base input settings. This dictates how many parameter combinations \emph{do not matter} to the recorded output response. This structural nonidentifiability, computed around a selected output response (one in a base setting) is a property of the system in a neighborhood of that setting, as long as the intrinsic dimensionality of the responses does not change when we perturb the base parameter values (\ie, as long as the QSSA approximation remains valid, see the discussion in Appendix~\ref{sec:selecting_base_values}). For our example, it was sufficient to perform \emph{linear} data processing of the optima by Principal Component Analysis. Indeed, the three redundant parameter combinations for the reference trajectory happen to live on a 3D hyperplane in full parameter space; this hyperplane contains $\sim\!99\%$ of the total variance of the 6D parameter vectors in the optimization set. In this example, it so happens that linear data analysis (PCA) is sufficient to determine the ``minimal response richness:'' the responses lie on a three-dimensional hyperplane in the ten-dimensional output space. In general (and, we expect, most often), PCA will suggest  \emph{more} than the truly minimal number of effective parameters to span the data, and nonlinear tools like DMaps would be required to find a minimal set.
	
	We already have our first result: a data-driven corroboration of the \emph{number} of effective parameter combinations. Three of them matter, and three of them do not, adding up to the correct total number of six full inputs. The reader may already have noticed that these structurally unidentifiable combinations \emph{are not global}; they are valid only for the reference trajectory. Beyond finding this number, we will also construct a global parameterization/foliation of the ``hypersurfaces that do not matter'' in the original input space. Even though only three-dimensional, they are impossible to visualize, leading to our introduction of a visualizable caricature below.
	
	\subsection{Effective Parameter Identification}\label{sec:parameter_identification}
	The leading non-harmonic eigenvectors, $\vect{\phi}$, computed for the transient data $\textbf{X}$ provide an intrinsic parameterization of this data set, \ie, a set of coordinates parameterizing the model manifold (see the discussion in Appendix~\ref{sec:diffusion_maps} for clarification of the term non-harmonics). However, they are not necessarily physically meaningful. In order to interpret them, the data scientist who knows their dimensionality can now ask a domain scientist to suggest a set of physically meaningful parameter combinations, $\kappa_i$, and try to quantitatively establish a one-to-one correspondence between the data-driven $\phi_i$ and the hypothesized meaningful $\kappa_i$. This approach to interpretability has been proposed and used in~\cite{frewen2011coarse,kattis2016modeling,meila2018regression,sonday2009coarse} for the case of data-driven effective \emph{variables}, and it can be extended, as we propose here, for data driven effective \emph{parameters}.
	
	In our case, Yeung \etal~have already provided us with good candidate analytical effective parameters $\vect{\kappa}=(\kappa_1,\kappa_2,\pi)\in {K}\subset\R^3$. We seek a (hopefully smooth) invertible mapping ${f:\Phi\to K}$ from the DMaps space to the space of analytical effective parameter values and back. This mapping is constructed through a ``slight twist'' on GH, which we call Double DMaps, explained in Appendix~\ref{sec:geometric_harmonics}. From the total 10,000 collected data points, we use 7,000 as training points and 3,000 as test points for our Double DMaps. We use the Inverse Function Theorem (IFT) described in Appendix~\ref{sec:explainability} to check that our data-driven effective parameters are indeed \emph{locally} one-to-one with the known analytical effective parameters (Equation~\ref{eqn:stas_effective_parameters}). We then use our Double DMaps GH to express the three theoretical effective parameters $\vect{\kappa}=(\kappa_1,\kappa_2,\pi)\in {K}\subset\R^3$ as (approximate) functions of our coordinates $\vect{\phi}$.
	
\begin{figure*}[htb]
		\centering
		\includegraphics[width=0.9\textwidth]{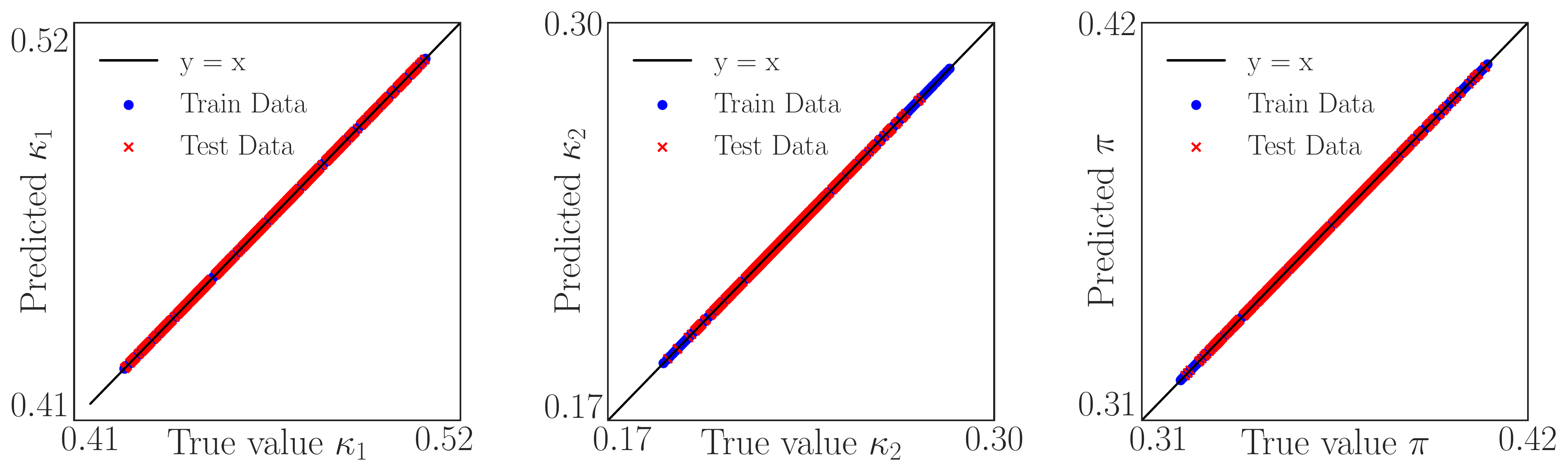}
		\caption{The three theoretical effective parameters predicted as a functions of the DMaps eigenvectors $\vect{\phi}$ with Double DMaps. Left $\kappa_1=f_1(\vect{\phi})$, middle $\kappa_2=f_2(\vect{\phi})$, right $\pi=f_3(\vect{\phi})$. Blue dots denote the training points (7,000 data points) and red crosses the test points (3,000 data points).}
		\label{fig:GH_interpolation}
	\end{figure*}

	An alternative realization of this map (data-driven effective to analytical effective) \emph{and its inverse} can also be constructed through the ``technology'' of neural networks: we used the data-driven effective parameters as inputs in a neural network whose outputs are the analytical effective parameters. Specifically, we used a five-layer, fully connected network with 30 nodes per layer and $\tanh$ activation functions, which we optimized via \texttt{ADAM} to achieve an MSE on the order of $10^{-6}$. Training this network provides an alternative realization of the mapping between the data-driven $\phi_i$ and the interpretable (here analytically obtainable) $\kappa_i$, the map $f:\Phi\to K$. We also obtained the inverse map, $f^{-1}:{K}\to\Phi$, by training a second neural network that implemented the same architecture and training scheme but with inputs and outputs switched. Instead of training two separate networks, one could combine the two networks into an autoencoder. Being able to construct the forward and the inverse mapping confirms the global one-to-one correspondence of the two sets on the data: the autoencoder would not be trainable otherwise.
	Figure~\ref{fig:GH_interpolation} plots the ground truth values of the three effective parameters against those interpolated with GH. 

\begin{figure*}[htb]
		\centering
		\includegraphics[width=0.9\textwidth]{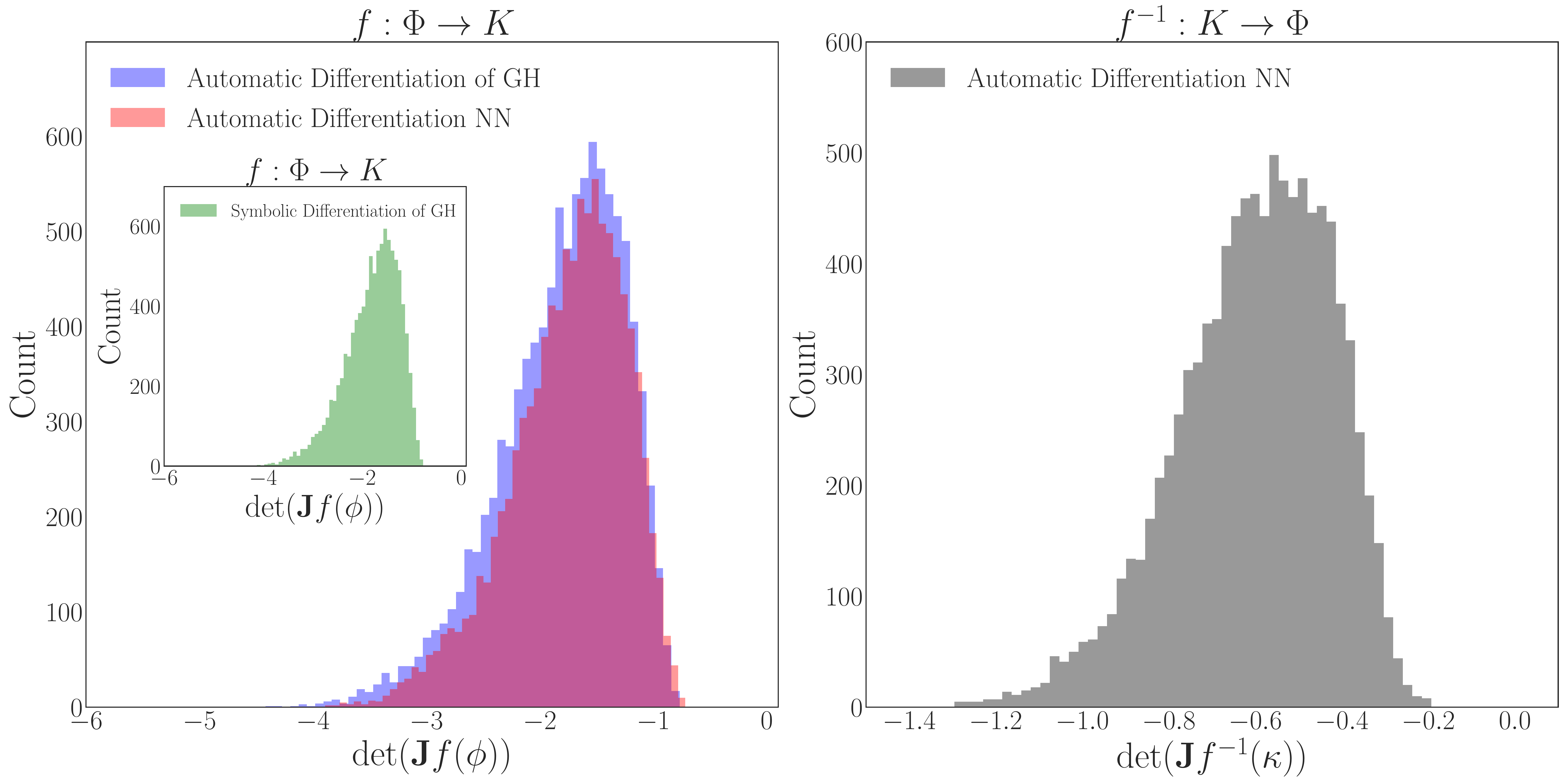}
		\caption{[Left] histograms of the determinant of the Jacobian, $\det\,\mathbf{J}f(\vect{\phi})$, computed on each observed data point with automatic and symbolic differentiation of GH and with automatic differentiation using a neural network. [Right] histogram of Jacobian determinants for the inverse function, $\det\,\mathbf{J}f^{-1}(\vect{\kappa})$, computed with a neural network.}
		\label{fig:Jacobian_determinants}
	\end{figure*}

	To establish that this map $f:{\Phi}\to{K}$ is invertible, we first confirm that the determinant of its $3\times 3$ Jacobian matrix is bounded away from zero for all points in our data set. By construction, $f$ is continuously differentiable, so the IFT guarantees local invertibility in a neighborhood of any point $\vect{\phi}\in\Phi$ where the Jacobian matrix
	\begin{equation*}
		\mathbf{J}f(\vect{\phi})=\begin{bmatrix}\partial\kappa_1/\partial\phi_1 & \partial\kappa_1/\partial\phi_3 & \partial\kappa_1/\partial\phi_9 \\ \partial\kappa_2/\partial\phi_1 & \partial\kappa_2/\partial\phi_3 & \partial\kappa_2/\partial\phi_9 \\ \partial\pi/\partial\phi_1 & \partial\pi/\partial\phi_3 & \partial\pi/\partial\phi_9\end{bmatrix}
	\end{equation*}
	is nonsingular. In Figure~\ref{fig:Jacobian_determinants}, we illustrate that $\det\mathbf{J}f(\vect{\phi})$ is bounded away from zero on our complete data set of $10^4$ points. Furthermore, our success in training the decoder component indicates that $f:\Phi\to\vect{K}$ is is globally invertible over our data and that our computed data-driven effective parameters are indeed one-to-one with the proposed theoretical ones, Equation~\ref{eqn:stas_effective_parameters}.
	
\begin{figure*}[htb]
		\centering
		\includegraphics[width=0.8\textwidth]{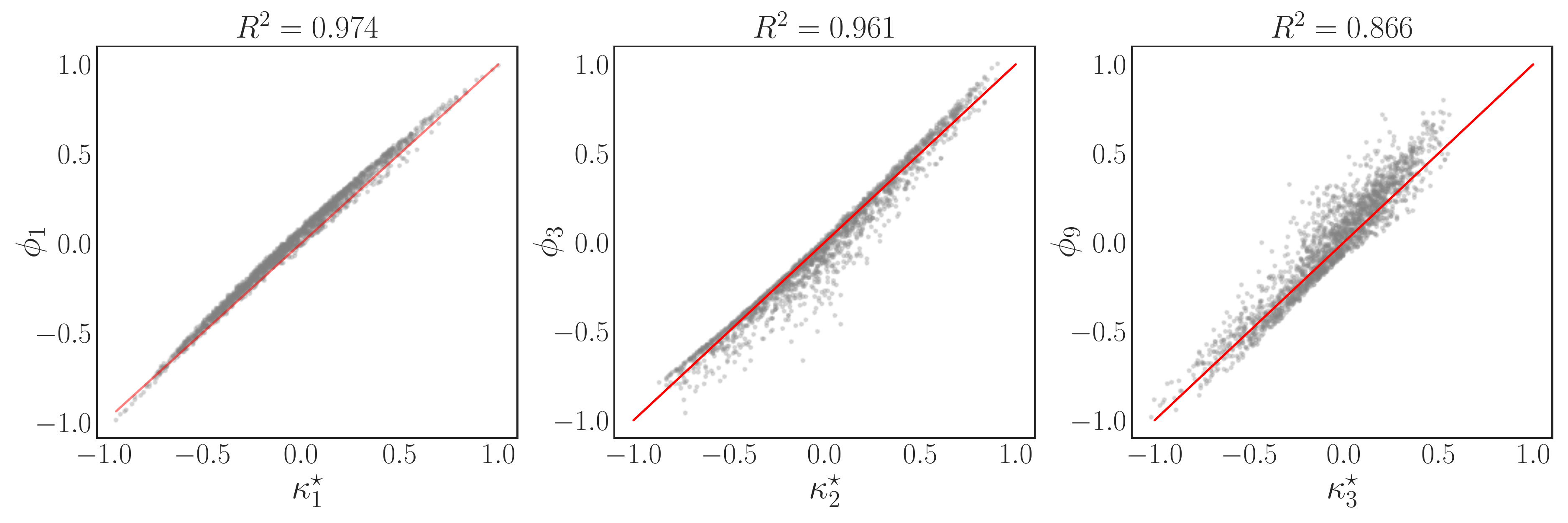}
		\caption{The three eigenvectors, $\vect{\phi}$ are fitted as functions of the original parameters, $\vect{p}$, through a symbolic regression algorithm. Entries of $\vect{\phi}$ and $\vect{p}$ were rescaled in the range $[-1,1]$. Expressions for the $\kappa_i^\star$ are provided in Equation~\ref{eqn:symbolic_regression}.} 
		\label{fig:Symbolic_Regression2}
	\end{figure*}

	The effective parameters proposed in~\cite{yeung2020inference}, were obtained by applying the QSSA to the full model. Simply by rearranging and simplifying the terms in Equation~\ref{eqn:stas_effective_parameters}, we could derive another equally plausible triplet of effective parameters: 
	\begin{align}
		\label{eqn:alt_effective_parameters}
		\kappa_1' & =\cE\frac{\kf1\kcat1}{\kr1+\kcat1}\ & \kappa_2' & =\frac{k_4}{k_5k_6},\ & \kappa_3' & =\frac{k_4}{k_6}.
	\end{align}
	Which of the two triplets would a symbolic regression package (\eg, gplearn~\cite{gplearn}) select? We illustrate an answer graphically in Figure~\ref{fig:Symbolic_Regression2} and analytically in Equation~\ref{eqn:symbolic_regression}. Note that, when performing this regression, we rescaled both the original parameters and the DMaps coordinates to lie in the range $[-1,1]$, as suggested in the package documentation~\cite{gplearn}:
	\begin{align}
		\kappa_1^\star & =0.288(\kcat2-\kcat1+\kr2+\kf1)\,,\nonumber\\
		\kappa_2^\star & =0.455(\kf1-\kf2)\,,\label{eqn:symbolic_regression}\\
		\kappa_3^\star & =0.218(0.36\kf1^2-1.38\kf1\kr2-\kf2+\kcat2\nonumber\\
		& \qquad\qquad-\kf1-\kr2-0.436)\,.\nonumber 
	\end{align}
	As illustrated in Figure~\ref{fig:Symbolic_Regression2}, these simple linear or quadratic expressions of the original parameters $\vect{p}$ can fit the coordinates quite accurately. In our opinion, while they can be written down in terms of ``simple cognitive basis functions,'' (\ie, monomials) ultimately these symbolically regressed parameters are almost as mechanistically uninterpretable as our data-driven effective ones. 
	
	\subsection{Behavior Estimation}\label{sec:behavior_prediction}
	Our computational formulation also allows us to obtain a mapping from new values of the effective parameters to the corresponding system output behavior. Each analytical effective parameter $\kappa_i$ and each element of every observed behavior vector are functions over the intrinsic model manifold, which is parameterized by the data-driven effective parameters $\phi_i$. If we are given a new triplet of $\phi_i$, GH on our Double DMaps can recover any element of any observation vector. If, on the other hand, we are given a new triplet of $\kappa_i$, we need only locally invert the known $\kappa_i(\phi_j)$ functions to the data-driven effective parameters, and proceed as above to predict the corresponding dynamic behavior through GH. Alternatively, after a round of DMaps on the $\kappa_i$, we perform GH on these DMaps to interpolate any desirable element of the expected behavior vector as a function of the $\kappa_i$. 
	
\begin{figure*}[ht]
		\centering
		\includegraphics[width=0.9\textwidth]{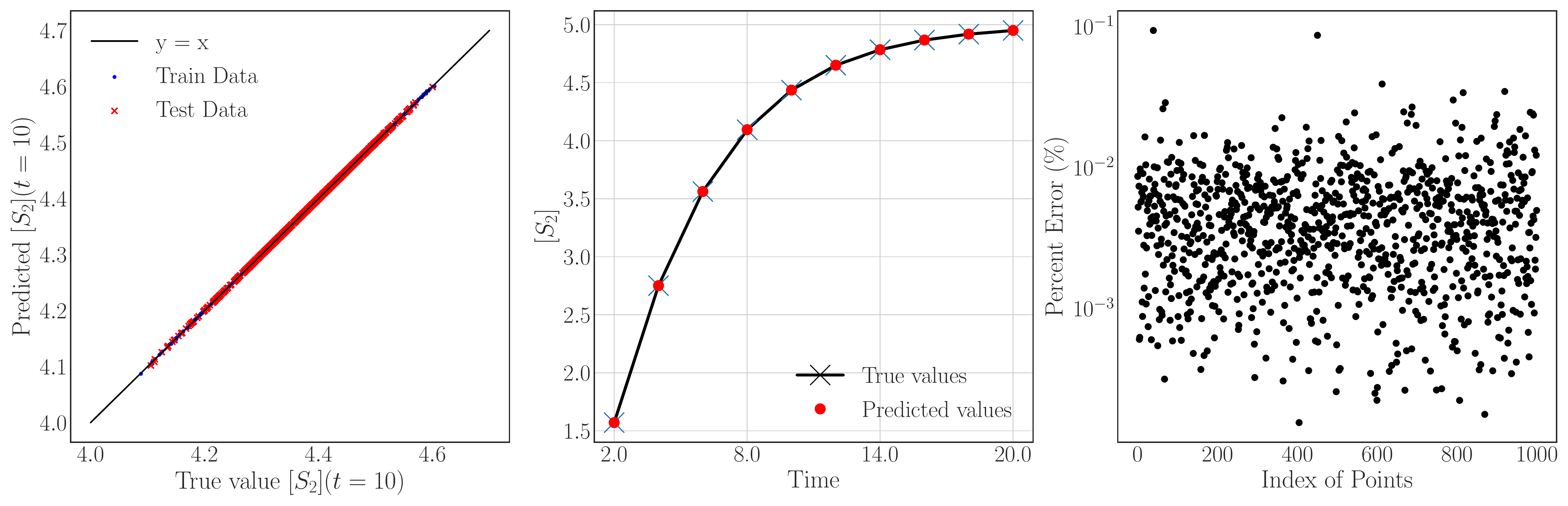}
		\caption{[Left] comparison of true and predicted values of the product concentration $\cStwo$ at $t=10$ with our scheme for 4,000 training and 1,000 test points. [Center] a reconstructed concentration profile of $\Stwo$ for a test point. With crosses are illustrated the true values, and with red points the predicted with Double DMaps point. [Right] the relative error for the 1,000 unseen behaviors.}
		\label{fig:GH_interpolation_for_time_series}
	\end{figure*}

	To implement this latter procedure, we generated 5,000 triplets of analytical effective parameters by perturbing uniformly within $\pm20\%$ of the nominal parameter values $(\kappa_1,\kappa_2,\pi)=(0.467,0.232,0.362)$, designating  4,000 as training and 1,000 as test points. We used this data set to learn the output concentration profiles for ten time steps of \ce{S2} with our Double DMaps GH scheme. Figure~\ref{fig:GH_interpolation_for_time_series} shows the true values of the concentrations against the predicted values with our scheme for $t=10$. Across all 1,000 test points for analytical effective parameter values, the relative prediction error does not exceed $0.1\%$.
	
	\subsection{Parameter Estimation}
	\label{sec:parameter_estimation}
\begin{figure*}[ht]
		\centering
		\includegraphics[width=\textwidth]{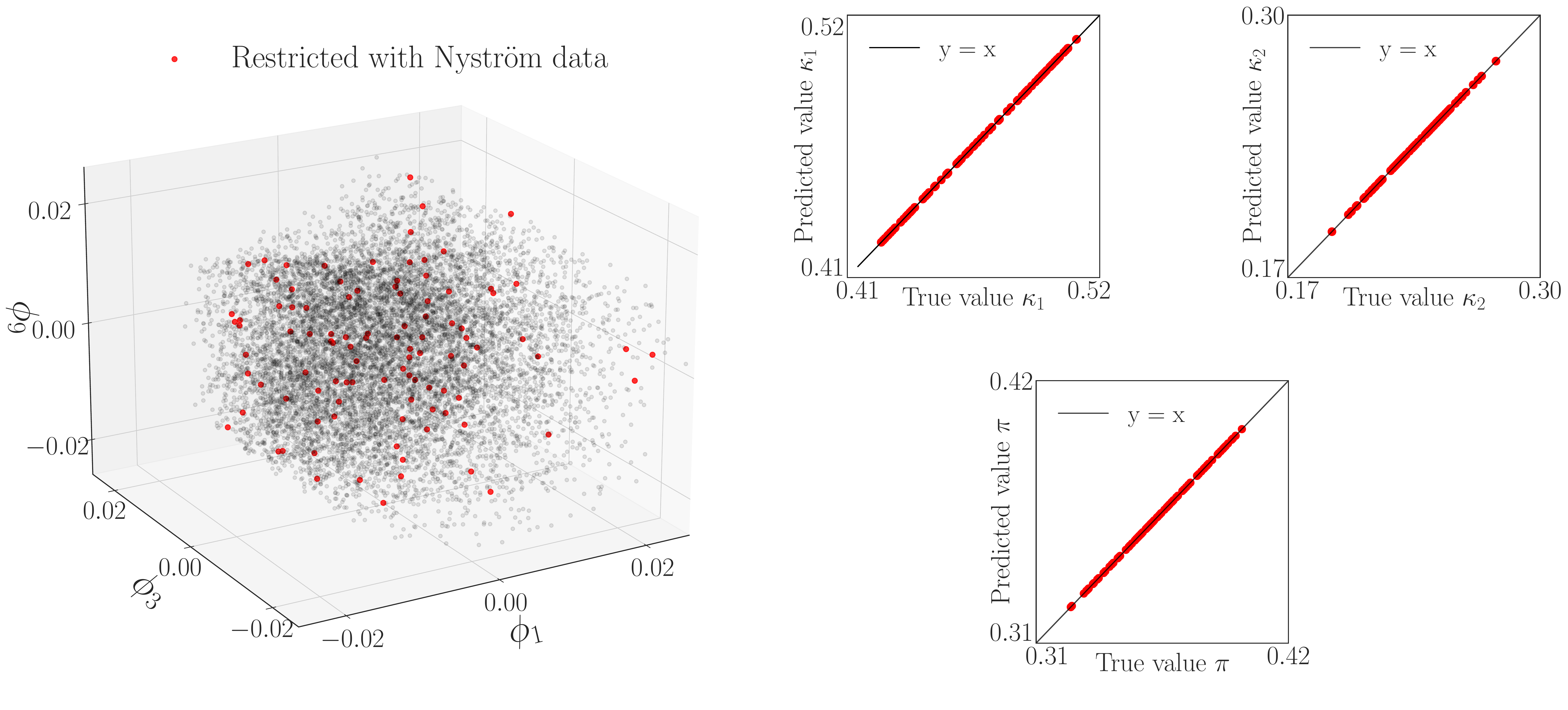}
		\caption{[Left] The unseen behaviors $\vect{f}(\vect{p}_{new})$ projected onto DMaps space via the Nystr\"om extension. [Right] For 100 unseen behaviors, the effective parameters $(\kappa_1,\kappa_2,\pi)$ are predicted with our Double DMaps scheme from previously-unseen behaviors $\vect{f}(\vect{p}_{new})$.}
		\label{fig:Nystrom_Prediction}
	\end{figure*}

	Even when the kinetic mechanism is known, parameter estimation is often challenging, due to measurement noise and differences in the timescales of individual reactions~\cite{yeung2020inference}. Estimating the parameters not through optimization, but through our data-driven scheme is straightforward from a technical standpoint. For previously unseen behaviors $\vect{f}(\vect{p}_\text{new}) = [\ce{S2}(t_1|\vect{p}_\text{new}),\ldots,\ce{S2}(t_{f}|\vect{p}_{new})]$, the Nystr\"om extension (described in Appendix~\ref{sec:nystrom_extension}) directly estimates the corresponding $\phi_i$ on the model manifold, from which we directly go to the effective parameters $\vect{\kappa}$ leading to this behavior through our Double DMaps version of GH (see Appendix~\ref{sec:geometric_harmonics}). Our approach performs this estimation in the minimal required dimensions---the intrinsic, data-driven ones---that jointly parameterize the observed behavior \emph{and} the meaningful input combinations that produce it. Figure~\ref{fig:Nystrom_Prediction} illustrates the projection of 100 previously unseen behaviors to the three-dimensional manifold through Nystr\"om extension and quantifies how well we can estimate the effective parameters for those unseen behaviors through our scheme.
	
	\subsection{On the Parameter Combinations that do not Matter}\label{sec:dont_matter_level_sets}
	Having identified a data-driven effective parameterization of the model and constructed data-driven maps from behavior to effective parameters and back, we now need to complete the task by mapping behavior to the original, full parameter set. 
	Clearly, this mapping is not one-to-one: for every observed behavior from the model, there exists \emph{an entire level set} of the original parameter space consistent with this behavior –and with a single set of meaningful parameter combination values. For the optimized data in Figure~\ref{fig:PCA_vs_DMAPs} we showed that an entire three-dimensional level set exists in the original parameter space, for a given output behavior (and so, for a given set of effective parameters/meaningful parameter combinations). But this does not identify the parameter combinations \emph{that do not matter}: those that, when they change along the level set, the effective parameter values as well as the resulting model behavior remain the same. In order to describe these level sets, we must employ a data-driven approach that allows us to detect the combinations of original parameters that do not affect the model output. This will disentangle the meaningful effective parameter combinations from the redundant ones. In Figure~\ref{fig:2D-illustration}, this disentangled parameterization was given by $\phi\equiv p_1p_2$ and $\psi\equiv p_1^2–p_2^2$.
	
	Notice that the level sets of these two types of original parameter combinations are conformal  everywhere. Moving $p_1$ and $p_2$ along the green level set does not change the model output, whereas moving them on the blue level set suffices to sample all possible output behaviors. In this way, the redundant parameter combinations allow us to construct the set of original, physical parameter values that are consistent with an observed behavior. Alternatively, holding them constant reduces the number of dimensions to be explored when optimizing the model behavior. Finally, after finding a behavior that optimizes a primary objective, the redundant parameter combinations help parameterize the search for an optimal \emph{secondary} objective---not a Pareto multiobjective but rather a \emph{lexicographic} optimization~\cite{arora2004introduction}. This disentanglement helps outline the nature of this subsequent lexicographic optimizationm and the dimensions available for it in parameter space. However, since the data are collected locally around the base point, our computation provides only a \textit{springboard} for further systematic exploration. A discussion of the systematic collection of additional data, parsimoniously extending the known ``patches" of the level sets is discussed in~\cite{chiavazzo2017intrinsic}. 
	
	\paragraph{A Visualizable Caricature.} The three-dimensional level sets of our working MSP example do not lend themselves to visualization. We therefore turn to a simpler kinetic model to illustrate these ideas and methods:
	\begin{align}
		\label{eqn:QSSA_Mechanism1}
		\ch{S_0 + E <>[\kfx][\krx] ES_0 ->[\kcatx] S_1 + E}\,,
	\end{align}
	where $\Szero$ and $\Sone$ are two different states of the substrate $\S$; $\E$ is the enzyme; and $\ESzero$ and $\ESone$ are complexes between the enzyme and the substrate. The differential equations can be found in Appendix~\ref{sec:toy_example}. We chose two base values of the original parameters $\kfx,\krx,\kcatx$ to work with. The first base value, $$\vect{k}_1=(\kfx,\krx,\kcatx)=(0.71,19,6700)\,,$$ gives a single effective parameter $\keffx\simeq\kfx$; in Appendix~\ref{sec:another-base-parameter-value}, we describe the discovery, through our manifold learning, of this single effective parameter and also the construction of its level sets. We choose to discuss here our results for the more interesting case of nominal parameters $$\vect{k_2} =(\kfx,\krx,\kcatx)=(0.97,7000,10000)\,.$$ In this regime QSSA, yields the single effective parameter:
	\begin{align}
		\label{eqn:k_eff2}
		\keffx=\Etot\,\frac{\kfx\kcatx}{\krx+\kcatx}\,,
	\end{align}
	where $\Etot$ is the total concentration of the enzyme.
	
\begin{figure}[ht]
		\centering
		\includegraphics[width=\textwidth]{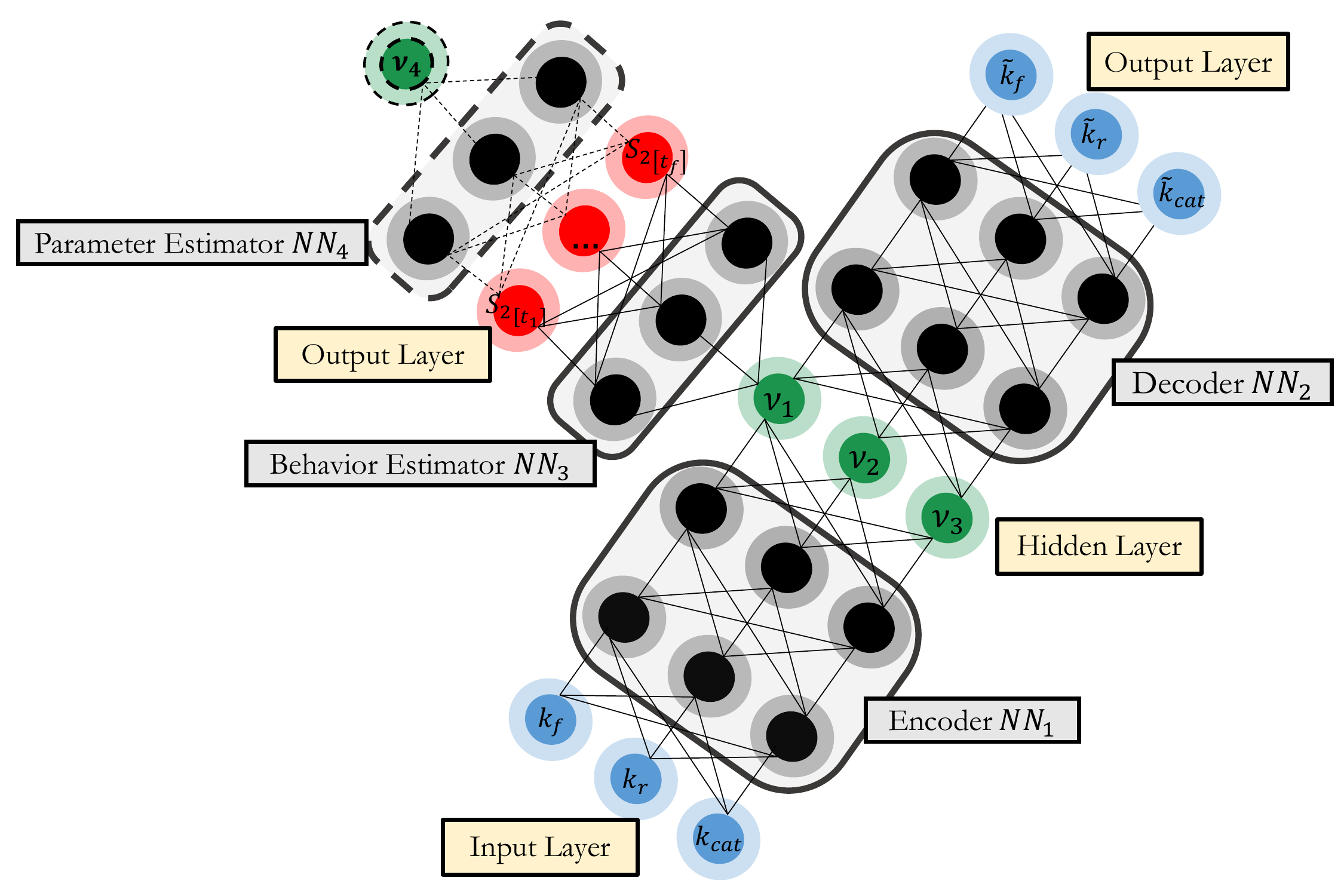}
		\caption{The proposed Y-shaped Conformal Autoencoder  consists of the following subnetworks: an Encoder ($\NN_1$), a Decoder ($\NN_2$), a Behavior Estimator ($\NN_3$), and possibly an additional Parameter Estimator ($\NN_4$). See Equation~\ref{eqn:named_neural_networks}.}
		\label{fig:Neural_Network_schematic}
	\end{figure}

	We generated 2,000 parameter vectors by sampling each entry uniformly within $\pm20\%$ of its nominal value. We collected output system behaviors for each parameter vector by integrating the model mechanism of Equation~\ref{eqn:QSSA_Mechanism1} from the reference initial condition $(\cSzero,\cE,\cSone,\cESone)=(5.0,0.66,0,0)$. The response is recorded every two seconds in time for five total points per trajectory. Our data-driven approach again detects that the output behavior of the system is intrinsically one-dimensional, and the new single effective parameter $\psi_1$ is one-to-one with our data-driven effective parameter $\keffx$, which is a combination of all three original parameters. The level sets of $\psi_1$ (or $\keffx$) are 2D curved surfaces (manifolds) in the original parameter space. In order to describe this level set, that is, discover the redundant parameter combinations,
	we introduced a \emph{Conformal Autoencoder} Y-shaped Neural Network architecture (see Figure~\ref{fig:Neural_Network_schematic}).
	
	Our Y-shaped Neural Network scheme consists of several connected subnetworks:
	\begin{align}
		\NN_1 & :(\kfx,\krx,\kcatx)\mapsto(\nu_1,\nu_2,\nu_3)\nonumber\\
		\NN_2 & :(\nu_1,\nu_2,\nu_3)\mapsto(\tilde{k}_{\text{f}},\tilde{k}_{\text{r}},\tilde{k}_{\text{cat}})\label{eqn:named_neural_networks}\\
		\NN_3 & :\nu_1\mapsto(\cStwo|_{t_1},\ldots,\cStwo|_{t_f})\nonumber\\
		\NN_4 & :(\cStwo|_{t_1},\ldots,\cStwo|_{t_f})\mapsto\nu_4\,.\nonumber
	\end{align}
	We used three multilayer perceptrons illustrated in Figure~\ref{fig:Neural_Network_schematic}:
	\begin{enumerate}
		\item an ``Encoder'' ($\NN_1$), which transforms the original parameters to a reparameterization, disentangling their meaningful combinations (one in the figure) and the redundant ones (two in the figure);
		\item a ``Decoder'' ($\NN_2$), which reconstructs the original parameters; and
		\item a ``Behavior Estimator'' ($\NN_3$), which maps the meaningful combination(s) to the observed output data.
	\end{enumerate}
	An additional ``Parameter Estimator'' ($\NN_4$) could be used to map observed behaviors back to the effective parameter(s) to ensure global invertibility.
	
	The key feature is the loss function, consisting of several parts. The obvious one is the successful reconstruction of the input original parameters (the ``Autoencoder'' part). Next comes the ability of $\NN_3$, whose input is the single effective parameter combination we seek, to reproduce the observed output; this forces $\nu_1$ to be one-to-one with the analytically known parameter $\keffx$. How many output measurements are necessary? Whitney's (and Takens') embedding theorems provide guarantees for $2n+1$ generic observations, when $n$ is the dimension of the model manifold~\cite{takens1981detecting}. Clearly, to build the architecture, we need to know in advance the number (here, one) of the required meaningful parameter combinations from the dimensionality of the model manifold. This number is the first quantity we compute with our output informed DMaps analysis of the transient system observations. The third necessary loss component comes from further imposing an \emph{orthogonality constraint} on the Conformal Autoencoder's latent coordinates $\vect{\nu}$:
	\begin{align*}
		\expval{\vect{d\nu}_i,\vect{d\nu}_j}=0\qquad\forall i\neq j\,,
	\end{align*}
	where $\vect{d\nu}_i$ indicates the vector of partial derivatives of the latent coordinate $\nu_i$ in terms of the input parameters ($\kfx,\krx,\kcatx$) and $\expval{\cdot,\cdot}$ indicates the inner product. This constraint is imposed using the automatic differentiation capabilities of the relevant code libraries and aims to disentangle what matters from what does not, making the architecture a ``Conformal Autoencoder.'' We explain the procedure used to train this Neural Network in Appendix~\ref{sec:Neural_Network}.
	
\begin{figure*}[ht]
		\centering
		\includegraphics[width=0.6\textwidth]{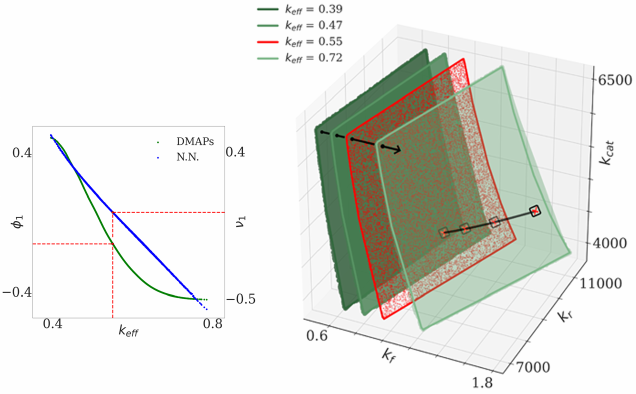}
		\includegraphics[width=0.7\textwidth]{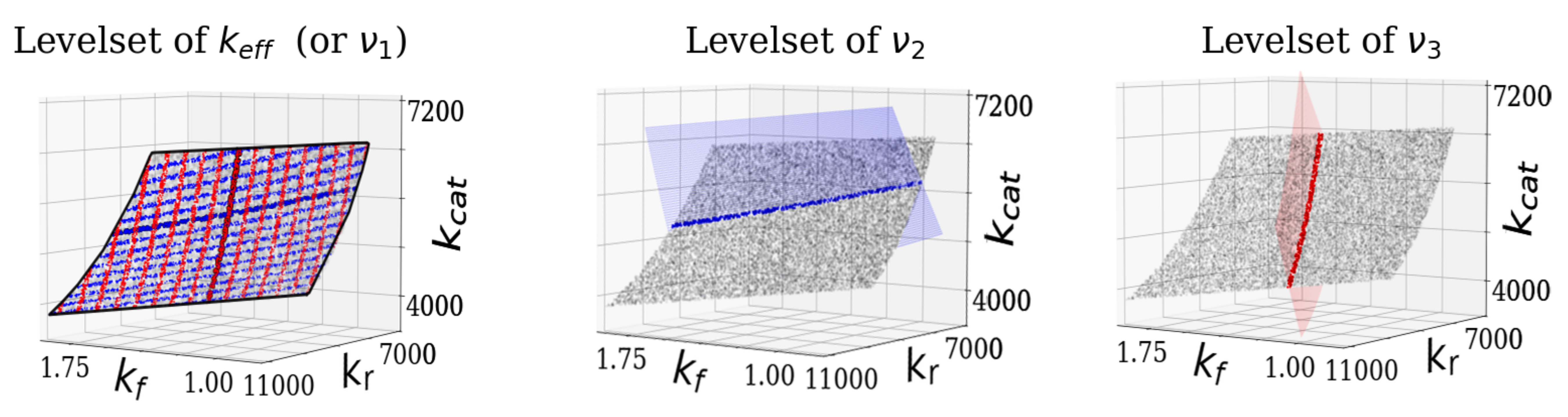}
		\caption{[Top left] the effective parameter $\keffx$ is one-to-one with the data-driven coordinate $\phi_1$, and also with the Neural Network effective variable $\nu_1$. [Top right] the level sets of constant behaviors, the level sets here are surfaces of the form $f(\kfx,\krx,\kcatx)=C$. A particular effective parameter (red point) corresponds to a level set (red surface) of the original parameters $(\kfx,\krx,\kcatx)$. [Bottom left] the same level set of $\keffx$ (equivalently, of $\nu_1$, since they are one-to-one), on which the conformal directions are colored as a grid of red and blue lines. [Bottom center] the intersection of the the level set of $\keffx$ with a level set of $\nu_2$. [Bottom right] the intersection of the level set of $\keffx$ with a level set of $\nu_3$.}
		\label{fig:figure2}
	\end{figure*}

	We thus discover a parameterization of the two redundant parameter combinations through $\nu_2$ and $\nu_3$. We also discover the Neural Network encoding of the effective parameter, $\nu_1$, which is one-to-one with both $k_\text{eff}$ and $\phi_1$ (see Figure~\ref{fig:figure2}). Our Double DMaps can easily approximate the estimation of $\nu_1$ from new, unobserved behavior. Figure~\ref{fig:figure2} shows representative (orthogonally) intersecting level sets of the three $\nu_i$, and the conformal grid of $\nu_2$, $\nu_3$ on a level set of the effective parameter $\nu_1$.
	
	This network can be used to encode a full set of initial parameter values to the effective parameter values that matter and through them to predicted behavior. More importantly, the already established path from new, unobserved behavior to the corresponding value of $\nu_1$, the effective parameter that matters, allows us to fix this value as an input to the Decoder $\NN_2$ and reproduce the level set of original parameters consistent with this new observed behavior by varying the values of $\nu_2$, $\nu_3$.
	
	\subsection{Jointly Smooth Function Extraction}\label{sec:JSF}
	We conclude this section by discussing how a kernel-based
	method called jointly smooth functions (JSFs), introduced by Dietrich \etal~\cite{dietrich2020spectral}, can be extended and used to disentangle input-output relations. Instead of a Neural Network architecture, the ``Jointly Smooth Functions''~\cite{dietrich2020spectral} approach, as its name suggests, could be used to find functions of the original parameters and functions of the output measurements that are \emph{jointly smooth} over the available data. Those \textit{jointly smooth} functions between the original parameters and the output are the effective parameters of the model in our case.
	
\begin{figure}[ht]
		\centering
		\includegraphics[width=\textwidth]{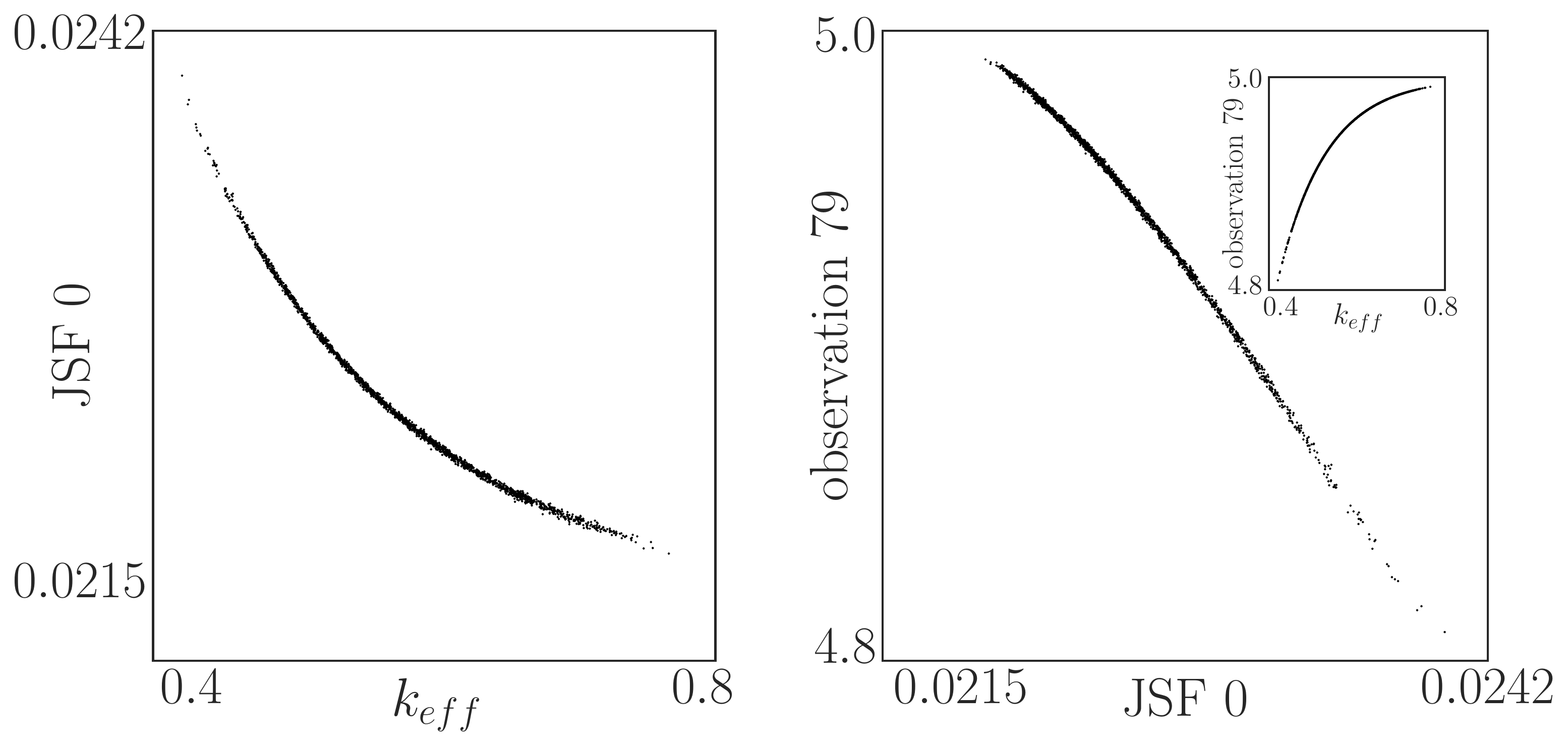}
		\caption{[Left] the first JSF for the second example, compared to the effective parameter $\keffx$. [Right] the first JSF is one-to-one with the observation $\vect{x}_1^{(79)}$.}
		\label{fig:jsf_example2}
	\end{figure}

	Figure~\ref{fig:jsf_example2} illustrates the results for our second, visualizable example. Two data sets are collected, containing 2,000 samples each. One consists of 20 time-delayed measurements of four output variable observations, $(\Szero,\Sone,\ESzero,\E)$, which we express as $\vect{x}_1\in\mathbb{R}^{80}$. The second contains the corresponding parameters $\vect{x}_2\in\mathbb{R}^3$. We use these two data sets as input to the JSF extraction pipeline (Algorithm in SI) and compute 25 such functions. The first JSF is one-to-one with the known effective parameter $\keffx$ (bottom left). We additionally plot an output (here one of the measurements, the 79\textsuperscript{th} one in time) that is also one-to-one with the first JSF (on the right). Note that, to test the robustness of the approach, the latter half of the output measurements were substituted with random noise uniformly distributed over the measurement range.
	
	In our work, we introduce an additional feature of the JSFs that allows to the computation of \emph{redundant} parameter combinations through the JSF approach; this is illustrated in Appendix~\ref{sec:JSF_app}.
	
	\section{Discussion}\label{sec:discussion}
	We have presented a systematic, data-driven approach for obtaining a meaningful reparameterization of parameter-dependent dynamical systems, disentangling the parameter combinations that matter to the output observations (temporal state measurements) from those that do not. The approach is generally applicable to the reparameterization of input-output relations.
	
	We used manifold learning techniques, including DMaps, to jointly parameterize the behaviors observed (the ``model manifold'') and the parameter combinations leading to them. We found the minimal number of meaningful parameter combinations (the effective parameters), expressed the outputs as functions of these effective parameters, and showed how to construct data-driven mappings from new effective parameters to the estimated outputs (prediction) and from new output observations back to effective parameters (estimation). It is worth mentioning that, in the case of noisy outputs, the DMaps parametrization will be robust to output noise as long as the scale parameter $\sqrt{\eps}$ remains larger than the amplitude of the noise~\cite{coifman2006diffusion}.
	
	Disentangling the parameter combinations that affect the output from those combinations that do not (the redundant parameter combinations) was obtained through a conformal autoencoder neural network. This allows us to now provide, for any observed behavior, not only the effective parameter values for it, but also \emph{the level set, in full input parameter space} consistent with this behavior. The capability of disentangling meaningful from redundant by enforcing conformality seems a promising research tool in tasks ranging from data-driven dimensional analysis to the exploration and construction of closures, and to the training of overparameterized neural networks.
	
	We briefly discuss the computational scalability of our approach. Generally, the ambient space dimension of the data influences the computational complexity less than the intrinsic dimension of the model manifold. \ie, the number of effective parameters. The detection of effective parameters in an intrinsically high-dimensional (say, five- or more dimensional) model manifold is less constrained by the scaling of our approach, but hinges on the large amount of data needed to sample the manifold well. Ambient space dimension, \ie, the number of given parameters (including redundant ones) as well as the number of observations, does not matter as much for the computational complexity of our approaches, since DMaps, GH, and JSFs are all based on pair-wise distance matrices that effectively ignore ambient dimension. The computational efficiency of the JSF approach is discussed in~\cite{dietrich2020spectral}. In general, kernel-based methods such as DMaps require more careful numerical implementations than Neural Network approaches, otherwise the number of data points becomes a bottleneck. Efficient algorithms that scale to millions of data points, even in high-dimensions, are available; see~\cite{dietrich2020spectral} and~\cite{shen2020scalability} for a discussion. Regarding memory, the Conformal Autoencoder network is less demanding than kernel-based approaches, because we can utilize mini-batching for training and highly parallelized software with efficient implementations is readily available. The analysis of the computational complexity of the network approach is much more involved than for kernel-based approaches, however, and out of the scope of this paper. Even convergence of the training is not clear, although some recent work hints on global convergence at least in controlled settings~\cite{jacot2018neural,rotskoff2018trainability}.
	
	It is interesting to consider the interplay of this approach with multi-objective optimization: if some input parameter combinations matter to a dominant objective, while others do not, we can, after a first round of optimization, exploit the redundant parameter combinations and optimize a second, ``subservient'' objective on optimal level sets of the first, dominant one. This is termed \emph{lexicographic} optimization and can also be related to ``lifelong learning.'' A conceptually simple example is the training of an overparameterized neural network to perform some task: the primary objective will be the accuracy of the prediction, while the ``subservient,'' secondary objective can be the pruning of the network for sparsity \textit{while remaining on the level set of successfully optimized predictions}. 
	
	Finally, we explored interpretability of our data-driven effective parameters through establishing bijections between them and candidate ``tuples'' of physical ones, that must come from domain experts. We also explored another simple approach to effective parameter interpretability by symbolically regressing the data driven effective parameters as functions of the input ones. 
	
	This work, creating mappings between parameters (in a sense, inputs to a dynamical system) to observed behavior (outputs) can be extended to create mappings between inputs and states, as well as mappings between states and outputs. We are exploring this direction towards data-driven balanced realizations. We expect that our level set parameterizations of the parameter sets that matter/do not matter (whether through Conformal NN or through JSF computations) may lead to useful extensions of the controllability and observability subspaces of linear theory. In this more general problem formulation, one can go beyond structurally unidentifiable inputs, and uncover spurious observations that are not system outputs (\eg, intrinsic sensor noise in our output observations)~\cite{talmon2019latent}. We are also exploring JSFs as a promising alternative kernel-based approach. Extracting the components of the inputs and outputs in the jointly smooth directions ``that matter'' can also help highlight those that do not. A key benefit is that, in addition to removing irrelevant input directions, this computation also removes output directions that are not influenced by the input (parameter) data, and provides a numerically stable and accurate approximation of the function space  over the space of the effective parameters.
	
	\section{Conclusion}
	We conclude by reiterating that, while the paper was focused on parameter nonidentifiabiity, in a context where the original model parameters function as ``inputs'' to the model, and the observed state time series are  the ``output'', our approach is generally applicable to data-driven (re)parametrization of more general input-output relations, with an eye towards disentangling meaningful inputs from redundant ones. Applicability of our current framework in an experimental setting
	involves (after selection of a reference set
	of conditions) the systematic local perturbation of all distinct experimental parameters/inputs; data mining on the response/output then leads to the  discovery of the meaningful and of the redundant parameter combinations. 
	
	\section*{Acknowledgments}
	The authors are grateful to Professors Mark Transtrum and Stas Shvartsman for helpful discussions. The authors thank the anonymous reviewers for their valuable suggestions.
	
	This work was partially supported by the U.S. Department of Energy (DOE), the Airforce Office of Scientific Research (AFOSR) and by the DARPA Atlas program.
	
	\pagebreak
	\bibliographystyle{plain}
	\bibliography{references}
	
	\pagebreak
	\appendix
	\section{Kinetic Models}\label{sec:kinetic_models}
	\subsection{The MSP Model of Yeung \emph{et al.}}\label{sec:yeung_model}
	We consider the dual phosphorylation of a substrate $\S$ by an enzyme $\E$, which is the illustrated mechanism in Equation~(\ref{eqn:MSP_mechanism}). The substrate can exist in any of three different states (phospostates): $\Szero$, $\Sone$ and $\Stwo$, where the index denotes how many times the substrate has been phosphorylated. Using elementary reaction kinetics~\cite{rawlings2002chemical}, we derive the following system of first-order differential equations to describe the evolution of the system in time:
	\begin{align}
		\frac{d\cSzero}{dt} & =-\kf1\cE\cSzero+\kr1\cESzero\,,\label{eqn:full_MSP_first}\\
		\frac{d\cESzero}{dt} & =\kf1\cE\cSzero-(\kf1+\kcat1)\cESzero\,,\\
		\frac{d\cESone}{dt} & =\kcat1\cESzero-(\kr2+\kcat2)\cESone+\kf2\cE\cSone\,,\\
		\frac{d\cSone}{dt} & =-\kf2\cE\cSone+\kr2\cESone\,,\\
		\frac{d\cStwo}{dt} & =\kcat2\cESone\,,\\
		\frac{d\cE}{dt} & =-\kf1\cE\cSzero+\kr1\cESzero-\kf2\cE\cSone+\kr2\cESone+\kcat2\cESone\,,\label{eqn:full_MSP_last}
	\end{align}
	The conservation laws for substrate and enzyme are given, respectively, by
	\begin{align}
		\Stot & =\left.\cSzero\right|_{t=0}=\cSzero+\cSone+\cStwo+\cESzero+\cESone\,,\label{eqn:full_MSP_conservation1}\\
		\Etot & =\left.\cE\right|_{t=0}=\cE+\cESzero+\cESone\,.\label{eqn:full_MSP_conservation2}
	\end{align}
	It is worth mentioning that $\Szero$ and $\Sone$ bind reversibly to the enzyme, which leads to complexes $\ESzero$ and $\ESone$, respectively. We assume that the experiment begins with all substrate in the $\Szero$ state and all enzyme molecules free. That is, at $t=0$, we have
	\begin{equation}
		\begin{bmatrix}
			\cSzero\\\cESzero\\\cESone\\\cSone\\\cStwo\\\cE
		\end{bmatrix}=\begin{bmatrix}
			\Stot\\0\\0\\0\\0\\\Etot
		\end{bmatrix}\,.
	\end{equation}
	Following the exposition of~\cite{yeung2020inference}, all concentrations are expressed in micromoles per liter, and the net production rate of each species has units of micromoles per liter per minute.
	
	\subsection{The Reduced MSP Model}\label{sec:reduced-msp}
	If the values of the rate constants place us in the regime where
	\begin{equation}
		\Stot\ll\frac{\kr1+\kcat1}{\kf1}\,,
	\end{equation}
	then we can use the QSSA to obtain the following system of three linear differential equations:
	\begin{align}
		\frac{d\cSzero}{dt} & =-\kappa_1\cSzero\,,\label{eqn:reduced_MSP_first}\\
		\frac{d\cSone}{dt} & =\kappa_1(1-\pi)\cSzero-\kappa_2\cSone\,,\\
		\frac{d\cStwo}{dt} & =\kappa_1\pi\cSzero+\kappa_2\cSone\,,\label{eqn:reduced_MSP_last}
	\end{align}
	where
	\begin{align}
		\label{eqn:effective_parameters}
		\kappa_1 & =\cE\,\frac{\kf1\kcat1}{\kr1+\kcat1}\,, & \kappa_2 & =\cE\,\frac{\kf2\kcat2}{\kr2+\kcat2}\,, & \pi & =\frac{\kcat2}{\kr2+\kcat2}
	\end{align}
	are the (analytical) effective parameters proposed in~\cite{yeung2020inference}, and the initial conditions at $t=0$ are $\cSzero=\Stot$ and $\cSone=\cStwo=0$.
	
	\subsection{A Toy Example}\label{sec:toy_example}
	The mechanism in Equation (5) of the main text is governed by the system of differential equations
	\begin{align}
		\frac{d\cSzero}{dt} & =-\kfx\cE\cSzero+\krx\cESone\,,\label{eqn:toy_model_first}\\
		\frac{d\cESzero}{dt} & =\kfx\cE\cSzero-\krx\cESzero-\kcatx\cESzero\,,\\
		\frac{d\cSone}{dt} & =\kcatx\cESzero\,,\label{eqn:toy_model_S1}\\
		\frac{d\cE}{dt} & =-\kfx\cE\cSzero+\krx\cESzero+\kcatx\,,
		ES_0\label{eqn:toy_model_last}
	\end{align}
	with conservation laws of substrate and enzyme, respectively, as
	\begin{align}
		\Stot & =\left.\cSzero\right|_{t=0}=\cSzero+\cSone+\cESzero\,,\\
		\Etot & =\left.\cE\right|_{t=0}=\cE+\cESzero\,.
	\end{align} 
	The QSSA for $\ESzero$ gives the following simplified expressions:
	\begin{align}
		\frac{d\cSzero}{dt} & =-\keffx\cE\cSzero\label{eqn:simplifed_S0}\,,\\
		\frac{d\cSone}{dt} & =\keffx\cE\cSzero\label{eqn:simplifed_S1}\,,
	\end{align}
	where
	\begin{equation}
		\keffx=\Etot\frac{\kfx \kcatx }{\krx+\kcatx}\,.
	\end{equation}
	If $\krx\ll\kcatx$ the effective parameter reduces further to $\keffx\approx\kfx$.
	
	\section{Methodology}\label{sec:methodology}
	\subsection{Diffusion Maps}\label{sec:diffusion_maps}
	Many techniques exist for parsimoniously describing low-dimensional data sampled from high-dimensional embedding spaces, including among others Isomap~\cite{tenenbaum2000global}, Local Linear Embedding~\cite{roweis2000nonlinear}, and Laplacian Eigenmaps~\cite{belkin2003laplacian} as well as diffusion maps (DMaps)~\cite{coifman2006diffusion}, which is our preferred approach here. In this section, we first explain the  algorithm in a more general way and then we illustrate how it applies to our parameter reduction problem.
	
	Given a data set, $\textbf{X}=\{\vect{x}_i\}_{i=1}^{N}$ with each $\vect{x}_i\in\R^{m}$, the first step of a DMaps algorithm is to construct a random walk on the data. This is achieved by means of an affinity matrix $\textbf{A}\in\R^{N\times N}$ that characterizes the likelihood of making a transition from point $\vect{x}_i$ to $\vect{x}_j$. The entries of $\textbf{A}$ are computed in terms of a kernel, typically the Gaussian kernel, which is defined as
	\begin{equation}
		A_{ij}=\exp\left(-\frac{\norm{\vect{x}_i-\vect{x}_j}^2}{2\eps}\right)\,, \label{eqn:Kernel}
	\end{equation}
	where $\norm{\cdot}$ denotes an appropriate norm for the observations~\cite{coifman2006diffusion,lafon2004diffusion}. In this paper, we will consider only the $\ell^2$ norm. The scale parameter $\eps>0$ regulates the rate of decay of the kernel: for small values of $\eps$, only points that are close to each other appear connected in $\textbf{A}$, since distant points will have $A_{ij}\approx0$.
	
	For the purposes of discovering a low-dimensional manifold $\mathcal{M}\subset\R^m$ and performing dimensionality reduction, we want to recover the geometry of the manifold. If the data points $\textbf{X}$ are non uniformly sampled on the manifold, then to compute the intrinsic dimensionality regardless of the sampling density an appropriate normalization of the affinity matrix must be performed. Define a diagonal matrix $\textbf{P}\in\R^{N\times N}$ with entries
	\begin{equation}
		P_{ii}=\sum^{N}_{j=1}A_{ij}\label{eqn:P_computation}
	\end{equation}
	and compute the normalized affinity matrix
	\begin{equation}
		{\widetilde{\textbf{A}}=\textbf{P}^{-\alpha}\textbf{A}\textbf{P}^{-\alpha}},
		\label{eqn:Normalization_density}
	\end{equation}
	where we choose $\alpha=0$ if assuming uniform sampling of the data, and $\alpha=1$ otherwise. The kernel matrix $\widetilde{\textbf{A}}$ is renormalized again by the diagonal matrix $\textbf{D}\in\R^{N\times N}$ to construct a row stochastic matrix $\textbf{W}$:
	\begin{equation}
		\label{eqn:Markov_Matrix}
		\textbf{W}=\textbf{D}^{-1}{\widetilde{\textbf{A}}}
	\end{equation}
	where $\textbf{D}$ is computed as:
	\begin{equation}
		D_{ii}=\sum^{N}_{j=1}\widetilde{A}_{ij}.
	\end{equation}
	Computing the eigendecomposition of $\textbf{W}$ and selecting the eigenvectors that parameterize independent directions (non-harmonic eignvectors) yields a non-linear parameterization of the original data set $\textbf{X}$. In our work this selection was achieved by applying the algorithm suggested in~\cite{dsilva2018parsimonious}. If the number of (non-harmonic) eigenvectors that provide this embedding is less than the number of the original dimensions of the data set, the algorithm achieves dimensionality reduction. Selecting the important eigenvectors is not as straightforward as in Principal Component Analysis (PCA), where one identifies the dimensionality of a given data set based on the energy that is captured in the leading singular vectors---often by finding a ``knee'' in the singular value plot, after which point a negligible fraction of the energy is contained in subsequent vectors. However, it can be achieved by sorting the eigenvectors, $\vect{\phi}_i$, based on their eigenvalues $\lambda_{i}$ and removing eigenvectors that can be represented as functions of the previous ones (harmonics)~\cite{dsilva2018parsimonious}. Those selected non-harmonic eigenvectors reveal the \textit{intrinsic} geometry of a given data set $\textbf{X}$ sampled from a manifold $\mathcal{M}$. In our work here, similar to~\cite{holiday2019manifold}, using aims to extract the \textit{intrinsic} parameters of a model. We note that, in the presence of noise, there may be no clear threshold by which to distinguish effective parameter combinations from non-effective ones; at some point the observer must decide what is ``too small to be considered,'' which is subjective (even when thoughtful) and therefore precarious.
	
	Two complementary approaches can be used to extract the effective parameters of the system that are relevant for the output, resp. those that are not. The effective parameters can in principle be discovered from observations of the model output; each data point then consists of a vector of measurements from time-series of the system behavior, \eg, $\vect{f}(\vect{p}_i)=[\vect{f}(t_1|\vect{p}_\text{i}),\ldots,\vect{f}(t_{f}|\vect{p}_i)]$, obtained for different combinations of parameter values $\vect{p}$. The affinity matrix in this context is computed as
	\begin{equation}
		\label{eqn:output_only_kernel}
		A_{ij}=\exp\left(-\frac{\norm{\vect{f}(\vect{p}_i)-\vect{f}(\vect{p}_j)}^2}{2\eps} \right)\,.
	\end{equation}
	The obtained  non-harmonic eigenvectors indicate how many parameters or combinations of parameters of the original (full) model are meaningful and give an embedding for those parameters. It is worth noting, however, that when the mapping from parameter space to the model manifold is noninvertible, different parameters may give  identical model responses (output multiplicity): $\vect{f}(\vect{p}_i)=\vect{f}(\vect{p}_j)$ with $\vect{p}_i\ne\vect{p}_j$. In that case, the simple output informed kernel fails~\cite{holiday2019manifold}. To circumvent this issue, Holiday \etal\cite{holiday2019manifold} proposed the use of a more informative kernel found in Lafon's Thesis~\cite{lafon2004diffusion}. In this latter case the affinity matrix is computed by taking into account both the inputs and the outputs but at different scales, as
	\begin{equation}
		\label{eqn:input_output_kernel}
		A_{ij}=\exp\left(-\frac{\norm{\vect{p}_i-\vect{p}_j}^2}{\eps^2}-\frac{\norm{\vect{f}(\vect{p}_i)-\vect{f}(\vect{p}_j)}^2}{\eps^{c}} \right)\,,
	\end{equation}
	where $c=4$ (for $\eps <1$) allows the disambiguation of inputs leading to the same output.
	
	The complementary approach is used to compute the number of ``non-meaningful'' parameters, that do not affect the output behavior of the model. Here a data set $\textbf{Y}=\{\vect{p}_i\}_{i=1}^N $sampled for a fixed behavior of the system (see Section~\ref{sec:parameter-reduction}) is needed. The identifiable effective parameters (assuming no output multiplicity) will then also be fixed; the unidentifiable combinations of parameters (those consistent with the same system behavior) may take entire continua of different values. Affinity matrix elements for pairs of points ($\vect{p}_i,\vect{p}_j$) in this data set are computed directly in the original parameter space:
	\begin{equation}
		\label{eqn:input_only_kernel}
		A_{ij}=\exp\left(-\frac{\norm{\vect{p}_i-\vect{p}_j}^2}{2\eps}\right).
	\end{equation}
	The two approaches are complementary: the first one aims to discover the dimensionality of the ``meaningful'' effective parameters, while the second approach addresses the dimensionality of the ones ``non-meaningful'' for the output. The total number of meaningful and non-meaningful parameters should then add up to the number of the original parameters of the model. 
	
	\subsection{Nystr\"om Extension}\label{sec:nystrom_extension}
	The Nystr\"om extension is a technique for finding numerical approximations to eigenfunction problems of the form~\cite{fowlkes2001efficient}:
	\begin{equation}
		\label{eqn:NystromIntegral}
		\int_a^bW(x_j,x_i)\phi(x_i)\,dx_i=\lambda\phi(x_j)
	\end{equation}
	\noindent
	In our framework, Nystr\"om is used as a simple extension tool to generate the DMaps coordinates $\vect{\phi}_{new}$ for new, previously unseen sample points $\vect{x}_{new}\notin\textbf{X}$. This interpolation scheme, \hbox {$f:\vect{x}_{new}\mapsto\vect{\phi}_{new}$}, requires recomputing the kernel that was used during the dimensionality reduction step (and applying the same normalizations) discussed in Section~\ref{sec:diffusion_maps}. The Nystr\"om extension formula reads:
	\begin{equation}
		\label{eqn:Nystrom_Expression}
		{\phi}_{\beta} (\vect{x}_{new}) = \frac{1}{\lambda_{\beta}} \sum_{i=1}^N {W}(\vect{x}_{new},\vect{x}_i)\phi_{\beta}(x_{i})
	\end{equation}
	where $\phi_{\beta}(x_i)$ is the $i$-th component of the $\beta$-th eigenvector ($\vect{\phi}_{\beta}$) and $\lambda_{\beta}$ is the $\beta$-th eigenvalue.
	
	\subsection{Double Diffusion Maps and their Geometric Harmonics}\label{sec:geometric_harmonics}
	Geometric harmonics (GH)~\cite{coifman2006geometric,lafon2004diffusion} is a scheme based on the Nystr\"om method~\cite{nystrom1929praktische}, \textit{traditionally} used for extending a function $f$ defined on a data set $\textbf{X}$ sampled from a manifold $\mathcal{M}$ for $\vect{x}_{new}\notin\textbf{X}$. 
	
	In our case, we mostly aim to extend functions defined \emph{not in the ambient space coordinates}, but on the discovered latent-reduced coordinates $\vect{\phi}$. Therefore, GH needs to be computed on only these few ``governing'' coordinates $\vect{\phi}$. Before we explain the algorithm, it is important to make clear why, without this additional step, the mapping from the reduced coordinates to any function defined on the ambient space would not be possible. We remind the reader that, during the first round of DMaps on $\textbf{X}$, we discovered the intrinsic dimensionality along with the corresponding set of a few variables $\vect{\phi}$. These new variables were obtained as eigenvectors of an eigendecomposition. Of course, they were not the \emph{only} eigenvectors computed; but they were necessary and sufficient eigenvectors to achieve the dimensionality reduction. All the other harmonic eigenvectors were ``discarded''~\cite{dsilva2018parsimonious}. Discarding those eigenvectors and keeping only the non-harmonic ones achieves the desired dimensionality reduction (and the corresponding reduced embedding) but is unable to accurately approximate a function on the manifold based on the reduced coordinates only: GH with only the governing eigenvectors gives a (possibly badly) truncated reconstruction of the function. If, however, we perform again DMaps on these few governing DMaps coordinates $\vect{\phi}$ (``Double DMaps''), and compute a new full set of eigenvectors, $\vect{\Psi}$, we obtain a full basis for expressing functions \emph{on the reduced manifold}---and therefore, functions on the original data. We mention again that it is not necessary to use the \emph{absolute minimal} number of Dmaps eigenvectors that parameterize the manifold; more than the minimal will, in principle, also work well for function reconstruction. Differentiating the approximated (via Double DMaps GH) function with respect to the governing DMaps coordinates (either symbolically or via automatic differentiation) is what will allow us to test, with the help of the Inverse Function Theorem (Section~\ref{sec:explainability}), the explainability of these coordinates in terms of physical parameters. In addition, it allows us to perform parameter estimation for new unseen behaviors.
	
	As in ``single'' DMaps, the first step for GH is to compute an affinity matrix:
	\begin{equation}
		A_{ij}=\exp\left(-\frac{\norm{\vect{\phi}_i-\vect{\phi}_j}^2}{2\eps}\right)\label{eqn:output_kernel}
	\end{equation}
	Since it is symmetric and positive semidefinite, this matrix $\textbf{A}$ has a set of orthonormal vectors  $\psi_0,\psi_1,\ldots,\psi_{N-1}$ and non-negative eigenvalues ($\sigma_0\geq\sigma_1\geq\cdots\geq\sigma_{N-1}\geq0$)~\cite{wendland2004scattered}. Those eigenvectors are used as a basis set onto which we project and subsequently extend the function of interest $f$.
	More precisely, for $\delta>0$ we consider the set of truncated eigenvalues $S_{\delta}=\{\alpha\,:\,\sigma_{\alpha}>\delta\sigma_{0}\}$. In this truncated set we project $f$ evaluated in some scatter points:
	\begin{equation} 
		f\mapsto P_{\delta}f=\sum_{\alpha\in{S_{\delta}}}\langle f,\psi_{\alpha}\rangle\psi_{\alpha}\,,\label{eqn:Geometric_Harmonics_Projection}
	\end{equation}
	where $\langle\cdot,\cdot\rangle$ is the inner product.The extension of $f$ for $\vect{\phi}_{new}\notin\Phi$ (or $\vect{x}_{new}\notin\textbf{X}$) is defined as:
	
	\begin{equation}
		\label{eqn:Geometric_Harmonics_Extension}
		(Ef)(\vect{\phi}_{new}) = \sum_{{\alpha\in{S_{\delta}}}} \langle f,\vect{\psi}_{\alpha}\rangle\Psi_{\alpha}(\vect{\phi}_{new})\,,
	\end{equation}
	where
	\begin{equation}
		\label{eqn:Geometric_Harmonics_functions}
		\Psi_{\alpha}(\vect{\phi}_{new}) = \sigma^{-1}_{\alpha}\sum_{i=1}^{m} {A}(\vect{\phi}_{new},\vect{\phi}_i)\psi_{\alpha}(\phi_i)
	\end{equation}
	and $\psi_{a}{(\phi_i)}$ is the $i$\textsuperscript{th} component of the DMaps eigenvector $\vect{\psi}_{\alpha}$. The function $\Psi_{\alpha}$ are the GH we use. It is worth noting that using a truncated set $S_{\delta}$ is important to circumvent the numerical instabilities that will arise in Equation~\ref{eqn:Geometric_Harmonics_functions} when $\sigma_i\to0$.
	
	Beyond enabling the extension of a function defined on $\textbf{X}$ (or $\vect{\Phi}$) GH can be used to approximate the gradient of the function in terms of the original variables (or the  variables). Symbolic differentiation of Equation~\ref{eqn:Geometric_Harmonics_functions} gives a closed form expression of the gradient of $f$ in term of the independent variables. Having this capability in our ``toolkit'' allows us to perform scientific computation without relying on, say, a finite difference scheme:
	\begin{equation}
		\label{eqn:Geometric_Harmonics_Gradient}
		\textbf{D}\Psi_{\alpha}(\vect{\phi}_{new}) = \sigma^{-1}_{\alpha}\sum_{i=1}^{m} -\frac{(\vect{\phi}_{new}-\vect{\phi}_i)}{\eps}{A}(\vect{\phi}_{new},\vect{\phi}_i)\psi_{\alpha}(\phi_{i})\,.
	\end{equation}
	We could also compute the gradient of $f$ with automatic differentiation of Equations~\ref{eqn:Geometric_Harmonics_Extension} and~\ref{eqn:Geometric_Harmonics_functions}.
	
	\subsection{Choosing Base Parameter Values, Representative Initial Conditions, and More}\label{sec:selecting_base_values}
	The computations we report are obtained by sampling the model response in a finite neighborhood of a single ``base point'' in parameter space, and for a single given set of ``reference'' initial conditions. Generically, in simulating an $n$-dimensional nonconservative dynamical system, if more than one attractors exist, their basins of attraction are also $n$-dimensional; perturbing a random initial condition within one basin will not affect the ultimate behavior, which will eventually approach the same attractor. The boundaries separating different basins are co-dimensional sets (so $n-1$ dimensional), so the points on them are much more ``rare'' (a set of measure zero) compared to points in any basin. In the same spirit, the reference initial condition choice will not generically affect the dimensionality of the model manifold, and therefore our estimation of the number of effective parameters. The initial conditions in whose neighborhood the model manifold dimension actually changes, we expect to be non-generic (also measure zero, \ie, of lower dimension than the generic ones). In that sense, choosing a reference initial condition randomly should be representative.
	
	Selecting the base point in parameter space, however, requires more discussion. Away from regimes where the QSSA leads to lower-dimensional behavior, any base point in a finite neighborhood will lead to the same model manifold dimensionality. Yet this would be different from the model manifold dimensionality observed at base points in regimes where the QSSA assumptions holds. So, while the precise base point is not important, \emph{the parameter space regime} in which it is chosen (and in which the model manifold dimensionality remains the same) does matter. Characterizing these different regimes, and their relations to each other, the model manifold and its boundaries, constitutes part of the Model Boundary Approximation Method~\cite{transtrum2014model}. For an illustrative study of transitions from parameter regimes with one model manifold dimensionality to parameter regimes with a different model manifold dimensionality see also~\cite{holiday2019manifold}. 
	
	Finally, it is worth mentioning that ``the scale of the observer'' (the units in which the measurements are recorded, and the time intervals allowed to elapse between successive measurements) may also very much affect the numerical determination of the dimensionality of the response. If, for example, the time intervals in our time series measurements are extremely small, it will appear that the solution simply does not (appreciably numerically) change, even when the base parameters are changing.
	
	\subsection{Explainability: Inverse Function Theorem}\label{sec:explainability}
	Consider a linear system of $n$ equations in $n$ variables, which may be written in full as
	\begin{align}
		a_{11}x_1+\cdots+a_{1n}{x_n} & =y_1\nonumber\\
		\vdots\hspace{8mm}\ddots\hspace{8mm}\vdots\hspace{5mm} & \,\hspace{5mm}\vdots\\
		a_{n1}x_1+\cdots+a_{nn}{x_n} & =y_n\nonumber
	\end{align}
	or succinctly as the matrix equation $\textbf{A}\vect{x}=\vect{y}$. This system has a unique solution $\vect{x}_\star=\textbf{A}^{-1}\vect{y}$ if and only if the matrix $\textbf{A}$ is invertible. For nonlinear systems of the form
	\begin{align}
		f_1(x_1,\ldots,x_n) & =y_1\nonumber\\
		\vdots\hspace{4mm}\ddots\hspace{4mm}\vdots\hspace{4mm} & \,\hspace{4mm}\vdots\qquad,\\
		f_n(x_1,\ldots,x_n) & =y_n\nonumber
	\end{align}
	however, we are generally limited to techniques that provide local information about possible solutions $\vect{x}$ to the system $f(\vect{x})=\vect{y}$. Suppose that $\vect{x}\in\R^n$ is such a solution and $f:\R^n\to\R^n$ is a differentiable function. The Inverse Function Theorem~\cite{marsden1993elementary} states that, if the Jacobian matrix
	\begin{equation}
		\label{eqn:jacobian}
		\mathbf{J}f(\vect{x})=\begin{bmatrix}
			\frac{\partial f_1}{\partial x_1} & \cdots & \frac{\partial f_1}{\partial x_n}\\
			\vdots & \ddots & \vdots\\
			\frac{\partial f_n}{\partial x_1} & \cdots & \frac{\partial f_n}{\partial x_n}
		\end{bmatrix}
	\end{equation}
	is invertible, then we can find neighborhoods of $\vect{x}$ and $\vect{y}$ for which an inverse function $f^{-1}:\R^n\to\R^n$ exists that specifies a unique (local) solution $\vect{x}_\star$ for any $\vect{y}_\star$ sufficiently close to $\vect{y}$.
	
	For data-driven applications, we are interested in demonstrating that there exists a globally one-to-one mapping between a set of inputs $\{\vect{x}_i\}$ and a set of outputs $\{\vect{y}_i\}$, without any knowledge of an analytical expression for the relationship between the two. If we can compute or approximate all first-order partial derivatives in Equation~\ref{eqn:jacobian}, then we can assess the invertibility of $f$ on the basis of the Jacobian's determinant at each input. A square matrix is invertible if and only if its determinant is nonzero, so finding that $\det\mathbf{J}f(\vect{x}_i)$ takes values of a single sign on our input set suggests that the input-output mapping is a one-to-one relationship.
	However, we must also consider that it is possible for a function to be locally invertible everywhere but not globally invertible. Thus, we must also check the data to ensure that the only pairs of observations with similar outputs, $\norm{\vect{y}_i-\vect{y}_j}\approx0$ also have similar inputs, $\norm{\vect{x}_i-\vect{x}_j}\approx0$.
	
	\subsection{Determining Level Sets in Practice}\label{sec:determining_level_sets}
	Consider a dynamic model involving $m$ physical parameters. We observe $n$ output quantities of this model at a fixed parameter vector $\vect{p}$:
	\begin{equation}
		\vect{F}:\R^m\to\R^n:\vect{p}\mapsto\begin{bmatrix}
			F_1(p_1,\ldots,p_m)\\
			\vdots\\
			F_n(p_1,\ldots,p_m)
		\end{bmatrix}\,.
	\end{equation}
	Suppose, however, that the dependence of $\vect{F}$ on $\vect{p}$ can be reduced to $d<m$ effective parameters,
	\begin{equation}
		\vect{q}=\begin{bmatrix}
			\phi_1(p_1,\ldots,p_m)\\
			\vdots\\
			\phi_d(p_1,\ldots,p_m)
		\end{bmatrix}\,,
	\end{equation}
	such that the underlying behavior $f:\R^d\to\R^n$ satisfies $\vect{F}(\vect{p})=\vect{f}(\vect{q})$. Applying the multivariate chain rule to $\vect{F}=\vect{f}\circ\vect{\phi}$, we have
	\begin{equation}
		JF(\vect{p})=[Jf(\vect{q})]\,[J\phi(\vect{p})]\in\R^{n\times m}\,,
	\end{equation}
	and, since $\phi:\R^m\to\R^d$, it follows that $\text{rk}[JF(\vect{p})]\leq\text{rk}[J\phi(\vect{p})]\leq d$.
	
	Let $\vect{p}\in\R^m$ be a vector of physical parameter values in which we have some interest and let $\vect{q}=\phi(\vect{p})$ and $\vect{y}=f(\vect{q})=F(\vect{p})$ be the corresponding effective parameters and outputs, respectively. We are interested in the $(m-d)$-dimensional level set of physical parameter values that produce the same output observations:
	\begin{equation}
		\mathcal{M}=\{\vect{v}\in\R^m\,|\,F(\vect{v})=\vect{y}\}\,.
	\end{equation}
	The tangent space to this manifold at $\vect{p}$ is given by the nullspace of the Jacobian:
	\begin{equation}
		\label{eqn:tangent_space}
		T_{\vect{p}}\mathcal{M}=\mathcal{N}(\mathbf{J}\vect{F}(\vect{p}))=\{\vect{v}\in\R^m\,|\,[\mathbf{J}\vect{F}(\vect{p})]\vect{v}=\vect{0}\}\,,
	\end{equation}
	which consists of the directions along which a local linear approximation of the output predicts no change. Given the Jacobian matrix $\mathbf{J}\vect{F}(\vect{p})$ (also called the Sensitivity Matrix~\cite{brouwer2018underlying}) at a point $\vect{p}$, we can obtain a basis for its nullspace from a Singular Value Decomposition (SVD)~\cite{horn2012matrix}:
	\begin{equation}
		\mathbf{J}F(\vect{p})=\mathbf{U}\mathbf{\Sigma} \mathbf{V}^\top\in\R^{n\times m}\,,
	\end{equation}
	in which $\mathbf{U}\in\R^{n\times n}$ and $\mathbf{V}\in\R^{m\times m}$ are orthogonal matrices and $\mathbf{\Sigma}\in\R^{n\times m}$ has all zero entries except on the main diagonal, where $\Sigma_{ii}=\sigma_i\geq0$. The nullspace $\mathcal{N}(\mathbf{J}F(\vect{p}))$ is spanned by the columns of $\mathbf{V}$ that correspond to those singular values $\sigma_i$ that equal zero. Alternatively, one could perform an eigendecomposition of the \emph{sensitivity Fisher Information matrix}, which is defined in~\cite{brouwer2018underlying} as $[\mathbf{J}F(\vect{p})]^\top\mathbf{J}F(\vect{p})\in\R^{m\times m}$ and corresponds to the expected value of the Fisher Information Matrix in the case of standard Gaussian measurement error.
	
	This approach provides the tangent space only at the point $\vect{p}$ for which we compute the Jacobian. The data-driven methods proposed in this paper provide much more than a local tangent space: the entire (global over our data) level set manifold; an orthogonal set of coordinates on it; as well as a completion of this set, through our redundant parameter combinations, to coordinates orthogonal over the entire parameter space.
	
	\subsection{The Conformal Autoencoder Network}\label{sec:Neural_Network}
	For the calculations included in this paper, all sub-networks of the Y-shaped conformal autoencoder network had the same specifications: five fully connected linear layers with 20 neurons each; the first four layers have a $\tanh(t)$ activation function and the last one is linear. Algorithm~\ref{alg:conformal-autoencoder} illustrates the training scheme used for our Conformal Autoencoder.
	
	\begin{algorithm}[p]
		\textbf{\underline{Input}}:Data {${\kfx,\krx,\kcatx}$} and output behaviors $\vb{S}$.\\
		\textbf{\underline{Output}:} The weights of the three neural networks $\qty{\theta_{NN_1}, \theta_{NN_2}, \theta_{NN_3}}$.\\
		For { $t=1,2,...,T$}
		\begin{enumerate}
			\item Predict:$$\mqty(\nu_1,\nu_2,\nu_3)=\NN_1(\kfx,\krx,\kcatx)$$
			\item $$\mqty(\tilde{k}_{\text{f}},\tilde{k}_{\text{r}},\tilde{k}_{\text{cat}})=\NN_2\mqty(\nu_1,\nu_2,\nu_3)$$
			\item Compute Autoencoder and Conformality Losses: $$ L_1 = \text{MSE}(\vect{\tilde{{k}}},\vect{{k}})+ \alpha \sum_{\qty{(i,j):j>i}}\text{MSE}(\expval{{\vect{d\nu}_i},\vect{d\nu}_j},{0})$$
			\item Update Weights (here we just show gradient descent) :\begin{gather*}
				\theta_{\NN_1}-={\eta}_1\grad_{\theta_{\NN_1}}L_1\\
				\theta_{\NN_2}-=\eta_2\grad_{\theta_{\NN_2}}L_1\\
			\end{gather*}
			\item $$\mqty(\nu_1,\nu_2,\nu_3)=\NN_1(\kfx,\krx,\kcatx)$$
			\item $$\vect{\tilde{S}}=\NN_3\mqty(\nu_1)$$
			\item Compute Behavior Estimator Loss: $$ L_2 = \text{MSE}(\vect{\tilde{S}},\vect{S})$$
			\item Update Weights (here we just show gradient descent):
			\begin{gather*}
				\theta_{\NN_1}-=\eta_1\grad_{\theta_{\NN_1}}L_2\\
				\theta_{\NN_3}-=\eta_3\grad_{\theta_{\NN_3}}L_2\\
			\end{gather*}
		\end{enumerate}
		\caption{Conformal Autoencoder Training: we used a hyperparameter value $\alpha=33$ to scale the relative importance of the orthogonality relation; $\vect{S}$ is the vector of true output behaviors; and $\tilde{\vect{S}}$ is the estimate from the network.}
		\label{alg:conformal-autoencoder}
		\hrulefill
	\end{algorithm}
	
	In the example presented in this work, we specifically used \texttt{ADAM} as the optimizer. The optimization process is heuristic: note that one ``epoch'' consists of two optimization steps, one updating the $(\NN_1,\NN_2)$ network which is an autoencoder, and one updating the $(\NN_1,\NN_3)$ network. This algorithm does not include the additional step of training $\NN_4$ of Figure~\ref{fig:Neural_Network_schematic}.
	Alternative formulations of the training protocol are, of course, possible. The structure of the architecture is more generally applicable,
	beyond the specific choices made here, and optimizing it is a topic of current research. 
	
\begin{figure}[htb]
		\centering
		\includegraphics[width=\textwidth]{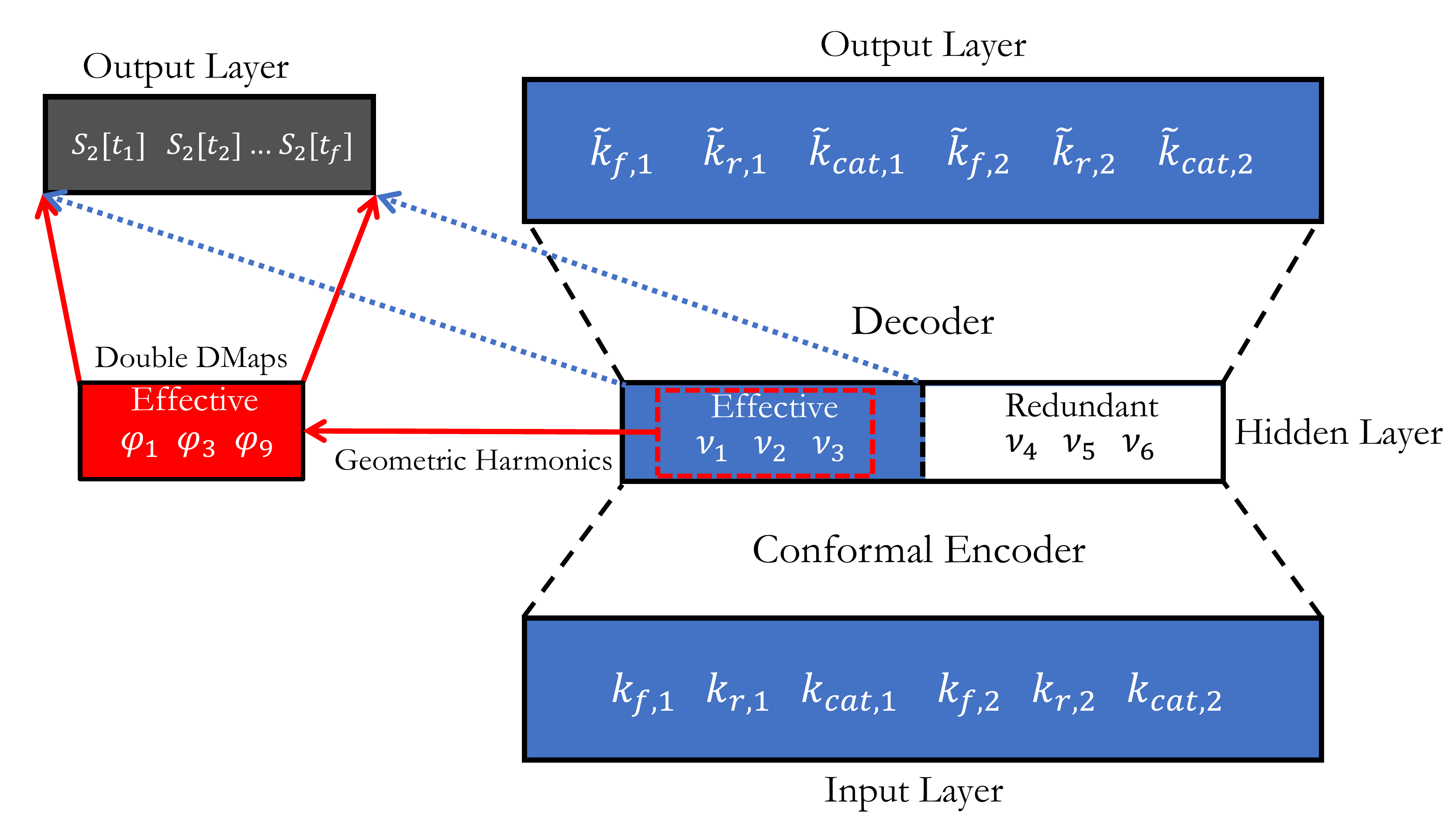}
		\caption{A schematic of the Y-shaped Conformal Autoencoder for the MSP example combined with our manifold learning scheme allows to make predictions for systems' behaviors given  new unseen full input vector $\vect{p}$.}
		\label{fig:Big_NN}
	\end{figure}

	In Figure~\ref{fig:Big_NN}, we illustrate how our Y-shaped conformal autoencoder will look for the MSP model discussed in Section~\ref{sec:msp_model}. In this case, we can directly map from the new full input vector to \emph{both} (a) the effective parameters of the Autoencoder ${\nu_1},{\nu_2},{\nu_3}$ and (b) from those latent descriptors to the estimated behavior. Alternatively, the path from $\vect{\nu}$ to the behavior can be implemented by using GH from the effective parameters of the Autoencoder to the DMaps coordinates and then from the DMaps coordinates with our Double DMaps GH scheme to the behaviors.
	
	\subsection{Jointly Smooth Functions}\label{sec:JSF_app}
	Jointly Smooth Functions (JSFs)~\cite{dietrich2020spectral} provide an alternative kernel-based pathway to obtaining effective parameters. The key idea for constructing JSFs between several (say, $K$) data sets, arising from different observations of the same phenomenon, is to define function spaces on all $K$ data sets separately, through eigenvectors of kernels that we will describe, and then use a singular value decomposition to find the ``common'' functions across these spaces. For details, see~\cite{dietrich2020spectral}. In our case, we have two data sets: the set of input/parameter settings for each simulation, and the set of output measurements for that input, and we will have to perform two eigendecompositions and a subsequent SVD. The ``common'' functions between input and output correspond to our effective parameters (meaningful parameter combinations, that affect the output); the ``uncommon'' functions between input and output correspond to our redundant parameter combinations, that do not affect the output.
	
	\begin{algorithm}[ht] 
		\hrulefill
		
		\textbf{\underline{Input}:} $K$ sets $\big\{ \vect{x}_{i}^{(1)},\vect{x}_{i}^{(2)},\dots\vect{x}_{i}^{(K)}\big\} _{i=1}^{N}$
		where $\vect{x}_{i}^{(k)}\in\mathbb{R}^{d_{k}}$.
		
		\textbf{\underline{Output}:} $M$ jointly smooth functions $\{ \vect{f}_{m}\in\mathbb{R}^{N}\} _{m=1}^{M}$.
		\begin{enumerate}
			\item For each observation set $\big\{ \vect{x}_{i}^{(k)}\big\} _{i=1}^{N}$
			compute the kernel:
			\[{K}_{k}(i,j)=\exp\left(-\frac{\big\Vert \vect{x}^{(k)}_{i}-\vect{x}^{(k)}_{j}\big\Vert^{2} }{2\sigma_{k}^{2}}\right)
			\]
			\item Compute $\textbf{W}_{k}\in\mathbb{R}^{N\times d}$, the first
			$d$ eigenvectors of $\textbf{K}_{k}$.
			\item Set $\textbf{W}=:\left[\textbf{W}_{1},\textbf{W}_{2},\dots,\textbf{W}_{K}\right]\in\mathbb{R}^{N\times Kd}$
			\item Compute the SVD decomposition: $\textbf{W}=\textbf{U} \textbf{$\Sigma$}\textbf{V}^{T}$
			\item Set $\vect{f}_{m}$ to be the $m$\textsuperscript{th} column of $\textbf{U}$.
		\end{enumerate}
		\caption{Jointly Smooth Functions from $K$ sets of observations.}
		\label{alg:multi}
		\hrulefill
	\end{algorithm}
	
\begin{figure}[ht]
		\centering
		\includegraphics[width=\textwidth]{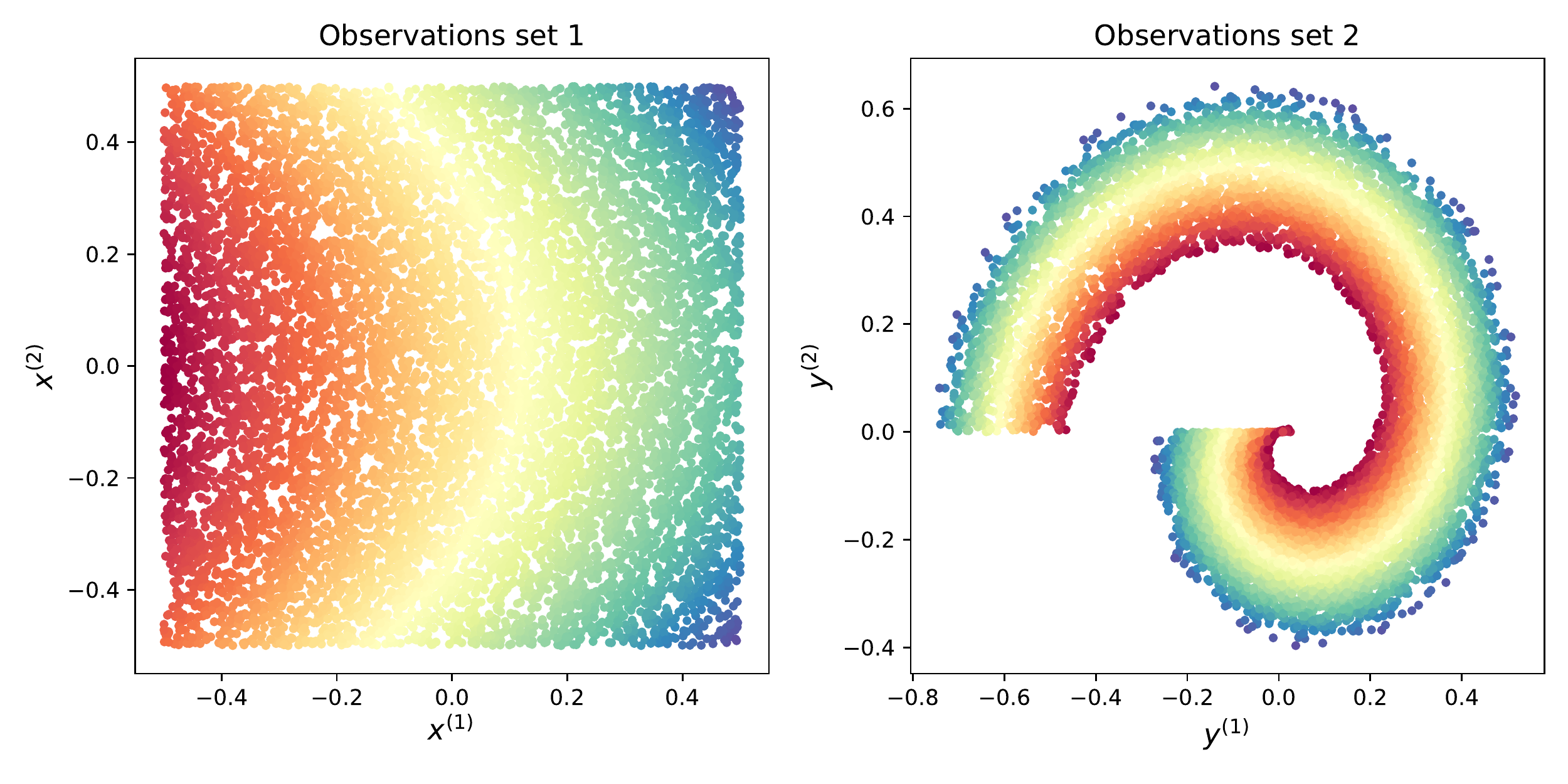}
		\caption{Two sets of measurements involving a common variable and ``set-specific'' uncommon variables, described in the text. Color indicates the ``common direction'' between the two data sets.}
		\label{fig:jsf_example_spiral}
	\end{figure}

	\paragraph{An Illustrative JSF Example} We illustrate through a toy example how the JSF algorithm discovers directions that are common between two data sets as well as directions that are ``uncommon'' between them, \ie, there are specific two the one or the other data set. For our application, the first data set consists of the parameter values and the second data set consists of the output measurements observed for these parameter values. For us the ``common'' directions between parameters and output observations correspond to our meaningful effective parameters; and the directions that are ``uncommon'' between parameters and output observations correspond to our redundant parameter combinations. Consider the random variable triplet $\left(a_{i},b_{i},c_{i}\right)\sim U\left[-0.5,0.5\right]^{3}$, iid and uniformly distributed. Define the ``common'' direction as $z_i=a_i+b_i^2$, and consider the first set of observations to be $\vect{x}_i=(a_i,b_i).$ The second set of observations is arranged on a spiral in $\mathbb{R}^2$ given by  
	\begin{equation}
		\label{eqn:spiral} \renewcommand*{\arraystretch}{1.1}
		\vect{y}_{i}=\left(\frac{c_{i}}{2}+\frac{z_{i}}{4}+\frac{1}{3}\right)\begin{bmatrix}\cos(2\pi c_{i})\\\sin(2\pi c_{i})\end{bmatrix}.
	\end{equation}
	The two sets are shown in Figure~\ref{fig:jsf_example_spiral}, with the common direction $z$ shown in color.
	
	Using Algorithm~\ref{alg:multi} to detect the variable that is ``common'' (jointly smooth), we then try to determine what is ``uncommon'' (sensor-specific) across our data-sets. In the example shown in Figure~\ref{fig:jsf_example_spiral}: how do we obtain a parameterization along the arclength of the spiral (uncommon between the data sets), as opposed to the parameterization across its width (common between the data sets)?
	
\begin{figure}[ht]
		\centering
		\includegraphics[width=\textwidth]{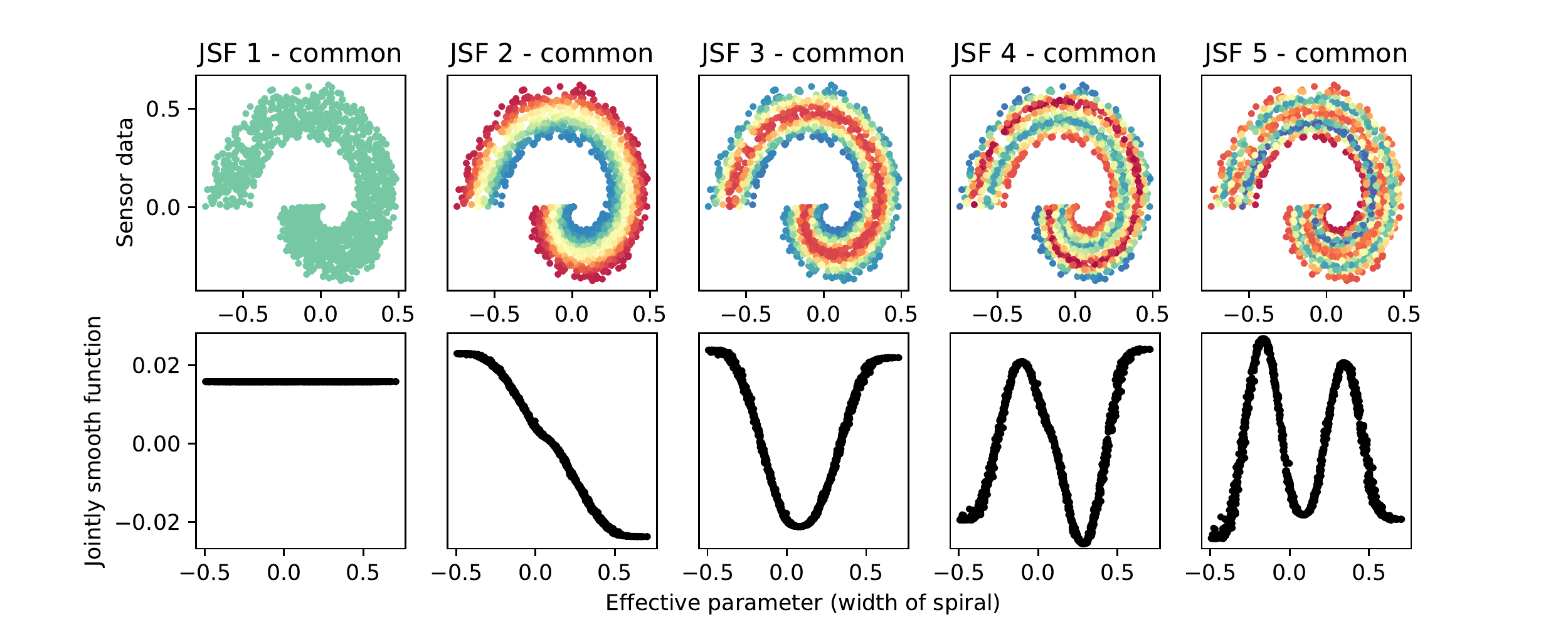}
		\includegraphics[width=\textwidth]{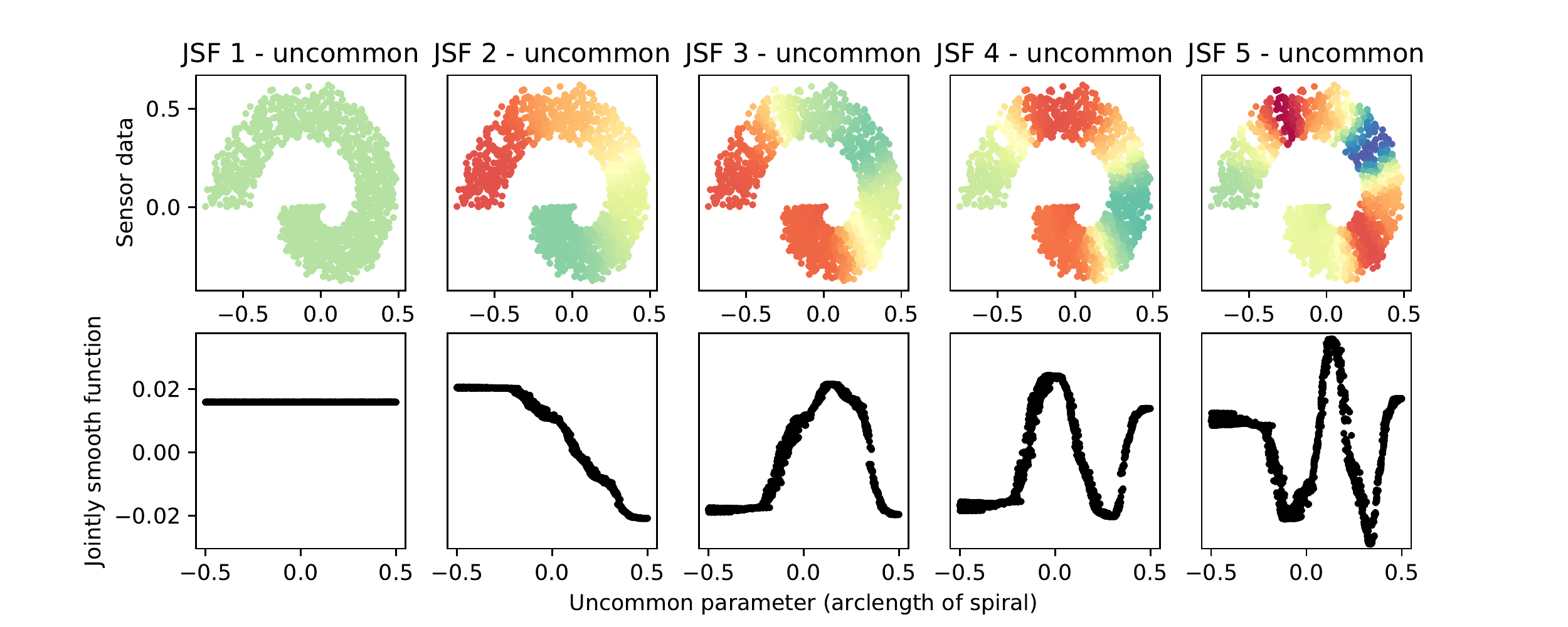}
		\caption{Common and uncommon functions extracted with the (extension of the) JSF algorithm on the spiral data set. Upon inspection, they can be rationalized as harmonics along the width (common) vs. harmonics along the arclength (the uncommon) directions.}
		\label{fig:jsf_spiral_common_vs_uncommon}
	\end{figure}

	Computationally identifying uncommon directions in the JSF framework can be performed as follows. After computing a set of JSFs $\vect{f}_{\text{JSF}}$ between the two data sets to obtain basis functions for the common subspace, we consider only one of the data sets (\eg, the spiral) and remove all of the common (jointly smooth) functions from the vector space spanned by the kernel eigenfunctions we computed in Algorithm~\ref{alg:multi}. Because the kernel eigenfunctions parameterize all functions on the manifold, what remains after removing common directions are functions that we expect to parameterize the uncommon directions. One issue with this approach is that we typically do not obtain enough JSFs to accurately span a large number of functions in the common directions. This would imply that removing only the small number of JSFs leaves too many common directions in the full space, and the uncommon eigendirections are still mixed with the common ones. To alleviate the problem of factoring out too few JSFs, in Algorithm~\ref{alg:jsf_uncommon_dmap}, we perform a different ``double'' process: by applying the Algorithm~\cite{coifman2006diffusion} to the few detected JSFs, we obtain a large number of smooth functions that span a larger portion of the function space on the common manifold. This larger number of common functions is then used to further enhance factoring out (removing the influence of) the common directions from the full space (Algorithm~\ref{alg:jsf_uncommon_dmap}, Step 2). When we apply Algorithm~\ref{alg:jsf_uncommon_dmap} to the spiral data, we obtain the results shown in Figure~\ref{fig:jsf_spiral_common_vs_uncommon}.
	
	\begin{algorithm}[ht]
		\hrulefill
		
		\textbf{\underline{Input}:} Full function space $\vect{f}_{\text{full}}\in\mathbb{R}^{N\times K}$, subspace to remove $\vect{f}_{\text{remove}}\in\mathbb{R}^{N\times R}$.
		
		\textbf{\underline{Output}:} Full space $\vect{f}_{\text{uncommon}}\in\mathbb{R}^{N\times K}$, with all functions only containing information in the uncommon directions.
		
		\textbf{\underline{Algorithm}:} 
		\begin{enumerate}
			\item Compute the projection of all functions on the functions to remove:
			\begin{equation*}
				c:=\vect{f}_{\text{full}}^T \vect{f}_{\text{remove}}\in\mathbb{R}^{K\times R}.
			\end{equation*}
			\item For $i=1,\dots,K$, select the $i$-th row of $c$, transpose, and multiply with the subspace to remove:
			\begin{equation*}
				\mathbb{R}^N\ni\vect{r}_i=\vect{f}_{\text{remove}} \underbrace{c_{i}^T}_{\in\mathbb{R}^{R}}\,.
			\end{equation*}
			\item For every function $\vect{f}_{\text{full},i}\in\mathbb{R}^{N}$, remove the contribution of all functions in the subspace:
			\begin{equation*}
				\vect{f}_{\text{uncommon},i}={\vect{f}_{\text{full},i}}-\vect{r}_i\,.
			\end{equation*}
		\end{enumerate}
		\hrulefill
		\caption{Obtaining uncommon directions in a function space.}
		\label{alg:jsf_uncommon}
	\end{algorithm}
	\begin{algorithm}[ht]
		\hrulefill
		
		\textbf{\underline{Input}:} Jointly smooth functions $\vect{f}_{\text{JSF}}\in\mathbb{R}^{N\times M}$, kernel eigenvectors $\vect{f}_{\text{kernel}}\in\mathbb{R}^{N\times K}$.
		
		\textbf{\underline{Output}:} Uncommon directions $\vect{f}_{\text{uncommon}}\in\mathbb{R}^{N\times M}$, with all functions only containing information in the uncommon directions.
		
		\textbf{\underline{Algorithm}:}
		\begin{enumerate}
			\item Apply DMaps to $\vect{f}_{\text{JSF}}$ to obtain a list of $R$ smooth functions $\Phi:=(\phi_1,\dots,\phi_R)\in\mathbb{R}^{N\times R}$, $R\gg M$, on the common directions, sorted by smoothness (DMaps eigenvalue).
			\item Apply algorithm~\ref{alg:jsf_uncommon} to the full space $\vect{f}_{\text{kernel}}$, removing the subspace $\Phi$, to obtain $\vect{f}_{\text{kernel, uncommon}}$.
			\item Apply the JSF algorithm to the following two data sets: (a) $\vect{f}_{\text{kernel, uncommon}}$ and (b) the observations $\vect{y}$ (the ones used to create $\vect{f}_{\text{kernel}}$). This creates a list of $M$ functions $\vect{f}_{\text{uncommon}}\in\mathbb{R}^{N\times M}$ that are ``jointly smooth'' between the uncommon functions obtained in step 2 and the original coordinates of the manifold, $\vect{y}$.
		\end{enumerate}
		\hrulefill
		\caption{Obtaining uncommon directions on a manifold.}
		\label{alg:jsf_uncommon_dmap}
	\end{algorithm}
	
	\paragraph{JSF Computations for Our Second Example.} 
	For our second example (Section~\ref{sec:dont_matter_level_sets}), we computed through Algorithm~\ref{alg:jsf_uncommon_dmap} also the redundant parameter combinations. The ``uncommon'' JSFs thus discovered for our second example  are colored with the Conformal Autoencoders' redundant coordinates. The figures support visually the one-to-one relationship between the two 
	descriptions.
	\begin{figure}[ht]
		\centering
		\includegraphics[width=\textwidth]{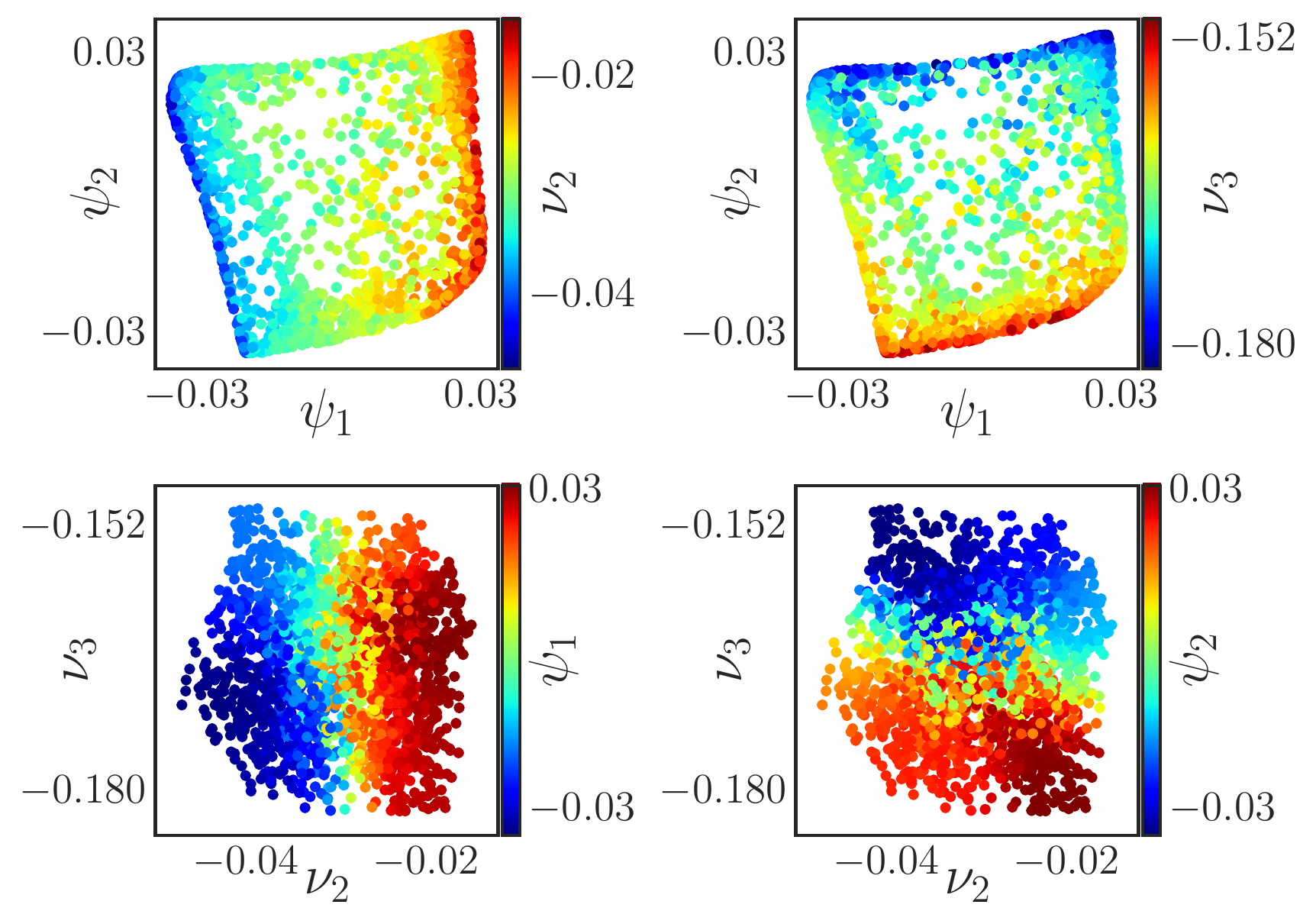}
		\caption{[Top] the uncommon (between parameters and output) JSFs, colored by the two redundant coordinates of the Conformal Autoencoder. [Bottom] the redundant coordinates of the Conformal Autoencoder, colored by the two uncommon JSFs.}
		\label{fig:jsf_example2-uncommon}
	\end{figure}
	
	\clearpage
	\section{Another Base Parameter Value Set for our Toy Example}\label{sec:another-base-parameter-value}
	For the second toy example we discussed in Section~\ref{sec:dont_matter_level_sets}, we also show what we find, with our scheme, around a different, even ``simpler'' base value, $\vect{k}_1 = (\kfx,\krx,\kcatx)=(0.71,19,6700)$. We follow the same algorithmic procedure,
	and compare our data driven effective parameter, obtained from the output informed DMaps, with the theoretical effective parameter based on QSSA. In this regime since $\kcatx \gg \krx$ the effective parameter based on the QSSA reads:
	\begin{align}
		\label{eq:effective_parameters}
		\keffx & =\Etot\frac{\kfx\kcatx}{\krx+\kcatx}\simeq \Etot\kfx 
	\end{align}
	
\begin{figure}[ht]
		\centering
		\includegraphics[width=0.9\textwidth]{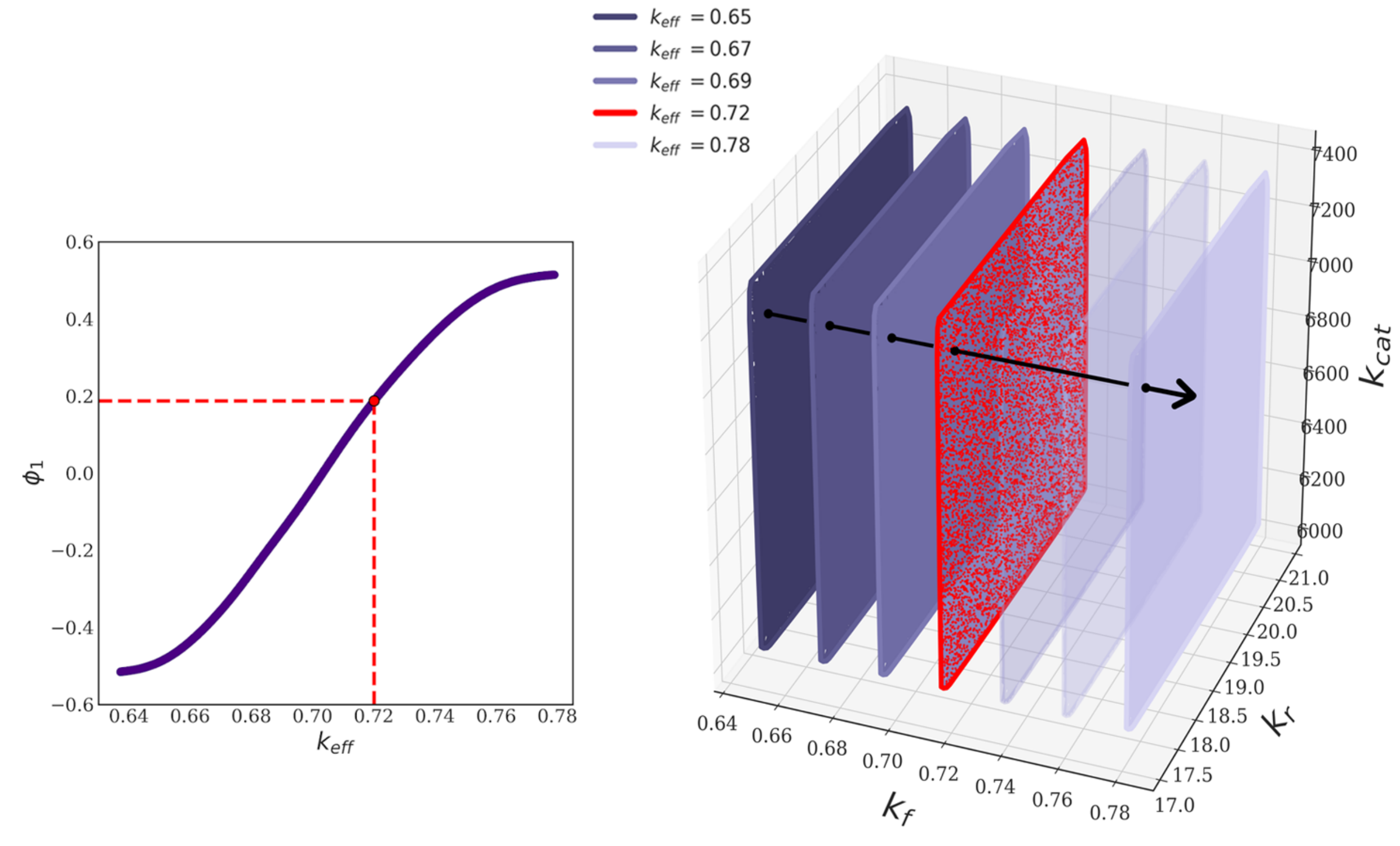}
		\caption{[Left] the data-driven coordinate $\phi$ is one-to-one with the effective parameter $\keffx$. [Right] levels sets of constant behaviors, the level sets are parallel to $\kcatx$, $\krx$. The red level set corresponds to the parameter combinations $\kcatx$, $\krx$ that give the behavior indicated with red point.}
		\label{fig:figure1}
	\end{figure}

	In this parameter regime, Figure~\ref{fig:figure1} demonstrates that our manifold learning approach discovers a single data driven effective input, $\phi_1$, the first nontrivial eigenvector of our output-informed Dmap computation. We confirm that this $\phi_1$ is one-to-one with the analytically (QSSA) obtainable effective parameter $\keffx$.
	In this regime, $\keffx$ is practically indistinguishable from the $\kfx$ input,
	and the level sets of $\phi_1$ (implicitly, the levels set of $\keffx$ and $\kfx$) are simply planes orthogonal to the $\kfx$ axis (parallel to the $\kcatx$ and $\krx$ axes in full input space).
	
	\section{Errors Fitting GH and NN}
	In this section, we report the errors computed for the different regression schemes mentioned in the main paper. The mean absolute percentage error for 3,000 tests points of the \text{forward} map, $f: \Phi \to K$, is reported for the interpolation schemes (Double DMaps GH and the Neural Network). Table~\ref{tab:TableMSE}. In Table~\ref{tab:Table-Errors-2}, the mean absolute percentage error for the \textit{inverse} map, $f^{-1}: K \to \Phi$ computed with the Neural is reported for the three DMaps coordinates. The mean absolute percent error for the prediction of effective parameters given unseen behaviors (Section~\ref{sec:parameter_estimation}) is shown in Table~\ref{tab:Table-Errors-3}.
	
	It is worth mentioning that the pre-trained Double DMaps GH for the mapping $f: \Phi \to K$ used for the  Nystr\"om formula was also used here. The reason we report also here the error is because of the use of the Nystr\"om extension formula for the restriction of the output observations to the reduced DMaps coordinates. 
	\begin{table}[ht]
		\centering
		\begin{tabular}{|c|c|c|c|}
			\hline
			Method & $\kappa_1$ & $\kappa_2$ & $\pi$ \\ \hline
			Double DMaps & $ 3.2 \times 10^{-3}  $ \% & $1.5 \times 10^{-4}$ \%  &$6.2\times 10^{-3} $ \% \\ \hline 
			Neural Network & $ 3.2 \times 10^{-2} $ \% & $4.0 \times 10^{-2}$ \%  &$3.4\times 10^{-2} $  \%\\ \hline
		\end{tabular}
		\caption{Mean absolute percentage error for the GH and the neural network interpolation scheme $f:\Phi\to K$.}
		\label{tab:TableMSE}
	\end{table}
	\begin{table}[ht]
		\centering
		\begin{tabular}{|c|c|c|c|}
			\hline
			Method & $\phi_1$ & $\phi_3$ & $\phi_9$ \\ \hline
			Neural Network & $ 2.9 \times 10^{-2} $ \% & $5.2 \times 10^{-2}$ \%  &$5.4\times 10^{-2} $  \%\\ \hline
		\end{tabular}
		\caption{Mean absolute percentage error with the neural network interpolation scheme \hbox{$f^{-1}:K\to\Phi$}.}
		\label{tab:Table-Errors-2}
	\end{table}
	\begin{table}[ht]
		\centering
		\begin{tabular}{|c|c|c|c|}
			\hline
			Method & $\kappa_1$ & $\kappa_2$ & $\pi$ \\ \hline
			Double DMaps & $ 3.0 \times 10^{-3} $ \% & $2.6 \times 10^{-4}$ \%  &$6.6\times 10^{-3} $  \%\\ \hline
		\end{tabular}
		\caption{Mean absolute percentage error with the Double DMaps scheme for the prediction of effective parameters $\vect{\kappa}$ for values of unseen behaviors.}
		\label{tab:Table-Errors-3}
	\end{table}
	
	\clearpage
	
	\section{Compartmental Models: A Textbook Nonidentifiability Example}\label{sec:compartmental_model}
	Compartmental models describe the exchange of matter or energy between different states~\cite{cole2020parameter}. This makes them suitable to applications in ecology, kinetics, separation processes and more~\cite{espana1978reduced, jacquez1988compartmental}. Such models consist of a system of coupled, first-order ordinary differential equations (ODEs). A \emph{linear} compartmental model with $n$ compartments can be written~\cite{cole2020parameter} in the form
	\begin{align}
		\vect{y}(t,\vect{p}) & =C(\vect{p})\,\vect{x}(t,\vect{p})\,,\nonumber\\
		\frac{\partial\vect{x}}{\partial t} & =A(\vect{p})\,\vect{x}(t,\vect{p})+B(\vect{p})\,\vect{u}(t)\,,\label{eqn:linear_compartmental_model}
	\end{align}
	where $A$, $B$, and $C$ are known matrix functions of a parameter vector $\vect{p}\in\R^d$; $\vect{u}(t)$ defines the input to the system; $\vect{x}\in\R^n$ are the internal system states; and $\vect{y}\in\R^m$ is a vector of observed quantities.
	
\begin{figure}[htb]
		\centering
		\resizebox{!}{2.4in}{
			\begin{tikzpicture}[node distance=1.5cm]
				\node (one) [compartment] {$1$};
				\node (input) [source, dashed, above of=one] {$u(t)$};
				\node (out1) [below of=one] {};
				\node (two) [compartment, right of=one, xshift=1cm] {$2$};
				\node (out2) [below of=two] {};
				\draw [arrow] (one) edge[bend left=15] node[midway,anchor=south]{$p_{12}$} (two);
				\draw [arrow] (two) edge[bend left=15] node[midway,anchor=north]{$p_{21}$} (one);
				\draw [arrow,dashed] (input) to (one);
				\draw [arrow] (one) edge node[midway,anchor=east]{$p_{10}$} (out1);
				\draw [arrow] (two) edge node[midway,anchor=west]{$p_{20}$} (out2);
			\end{tikzpicture}
		}
		\caption{Schematic of a compartmental model with four parameters, specifying the rates at which material is exchanged ($p_{12}$ and $p_{21}$) between two compartments and flows out of the system ($p_{10}$ and $p_{20}$). In the case of Equation~(\ref{eqn:linear_compartmental_model}), an impulse input $u(t)$ is initially supplied to the first compartment, the contents of which constitute our observations.}
		\label{fig:compartmental_model_diagram}
	\end{figure}
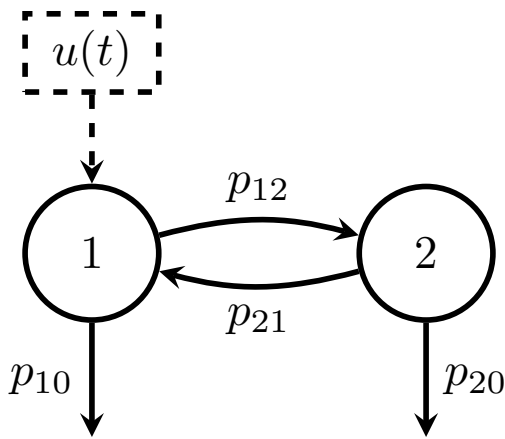

\begin{figure}[htbp]
		\centering
		\includegraphics[scale=1.0]{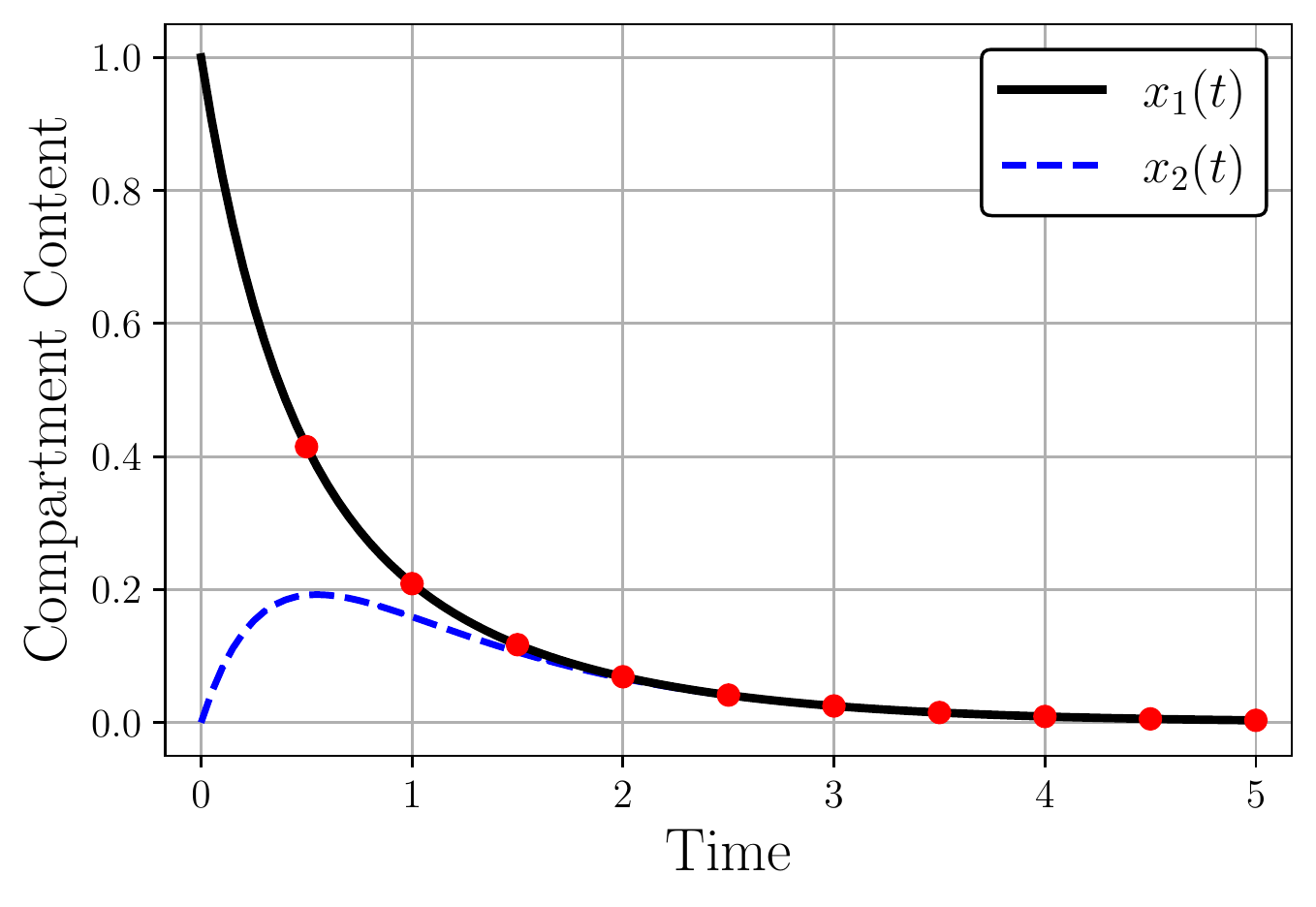}
		\caption{Dynamic behavior of the model in Equation~(\ref{eqn:linear_compartmental_model}) given reference parameter values $\tilde{\vect{p}}=(1,1,1,1)$. We observe only $x_1(t)$ at ten equally-spaced times, which are indicated by the red circles.}
		\label{fig:reference_trajectory}
	\end{figure}

	We borrow a textbook two-compartment model, illustrated in Figure~\ref{fig:compartmental_model_diagram}, for which we observe the scalar quantity $y(t)=x_1(t)$ given
	\begin{align}
		A & =\begin{bmatrix}-(p_{10}+p_{12}) & p_{21}\\p_{12} & -(p_{20}+p_{21})\end{bmatrix}\,, & B & =\begin{bmatrix}1\\0\end{bmatrix}\,,\nonumber\\
		C & =\begin{bmatrix}1 & 0\end{bmatrix}\,, & u(t) & =\delta(t)\,,\label{eqn:cole_model}
	\end{align}
	where $\delta(\cdot)$ represents a unit impulse at initial time $t=0$. Cole~\cite{cole2020parameter} demonstrates that this model is structurally nonidentifiable: its four parameters can be reduced to a set of three:
	\begin{equation}
		\vect{\beta}=\begin{bmatrix}\beta_1\\\beta_2\\\beta_3\end{bmatrix}=\begin{bmatrix}p_{10}+p_{12}\\p_{20}+p_{21}\\p_{12}\,p_{21}\end{bmatrix}\,.\label{eqn:cole_effective_parameters}
	\end{equation}
	We select as our base point $\tilde{\vect{p}}=(1,1,1,1)$ and take our output observation function to be
	\begin{equation}
		\vect{y}(\vect{p})=\begin{bmatrix}x_1(0.5,\vect{p}) & x_1(1.0,\vect{p}) & \cdots & x_1(5.0,\vect{p})\end{bmatrix}^\top\in\R^{10}\,,\label{eqn:compartmental_output_function}
	\end{equation}
	subject to fixed initial conditions $\vect{x}(0)=\begin{bmatrix}1 & 0\end{bmatrix}^\top$ (resulting from the impulse input $u$). The dynamic behavior starting at this base point, $\tilde{\vect{y}}=\vect{y}(\tilde{\vect{p}})$, is illustrated in Figure~\ref{fig:reference_trajectory}. We generate $N=5000$ parameter vectors $\{\vect{p}_i\}_{i=1}^N$ by independently and uniformly perturbing each component within $\pm10\%$ of its base value and record the corresponding output response history vectors $\{\vect{y}_i\}_{i=1}^N$. We used these first 5000 parameter settings to train our GH and NN models, while similarly generating an additional $N'=500$ as a test set, $\{\vect{p}_i'\}_{i=1}^{N'}$.
	
\begin{figure}[htbp]
		\centering
		\includegraphics[width=\textwidth]{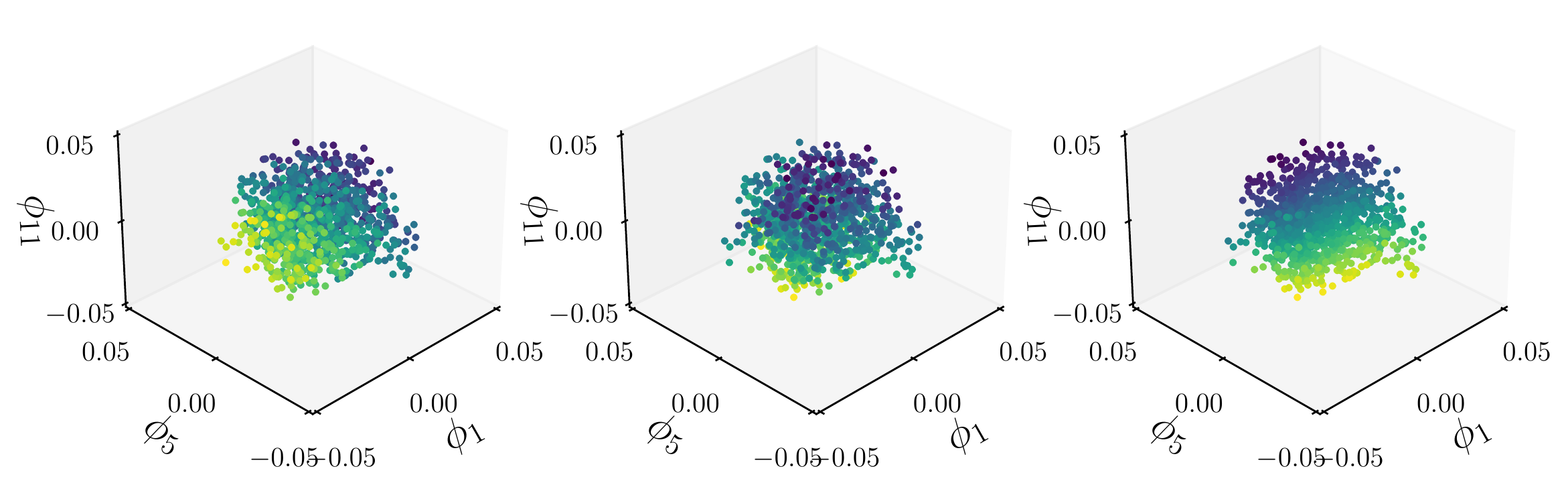}
		\caption{Our data-driven effective parameters, colored by theoretically proposed $\beta_1,\beta_2,\beta_3$ (left, center, right, respectively). The one-to-one correspondence between data-driven and theoretical parameters is visible.}
		\label{fig:dmaps_evecs_colored}
	\end{figure}

	We computed diffusion maps (DMaps) on the training output vectors and found that $(\phi_1,\phi_5,\phi_{11})$ form a reduced set of three effective parameters, while the other eigenvectors are functions of one or more of these. In Figure~\ref{fig:dmaps_evecs_colored}, we plot triplets of $\vect{\phi}=(\phi_1,\phi_5,\phi_{11})\in\R^3$ colored by the values of Cole's proposed effective parameters, defined in Equation~(\ref{eqn:cole_effective_parameters}). Visual inspection suggests that there is a bijective map, $h:\vect{\phi}\mapsto\vect{\beta}$, between the two sets of coordinates.
	
\begin{figure}[htb]
		\centering
		\begin{subfigure}{0.48\textwidth}
			\centering
			\includegraphics[width=\linewidth]{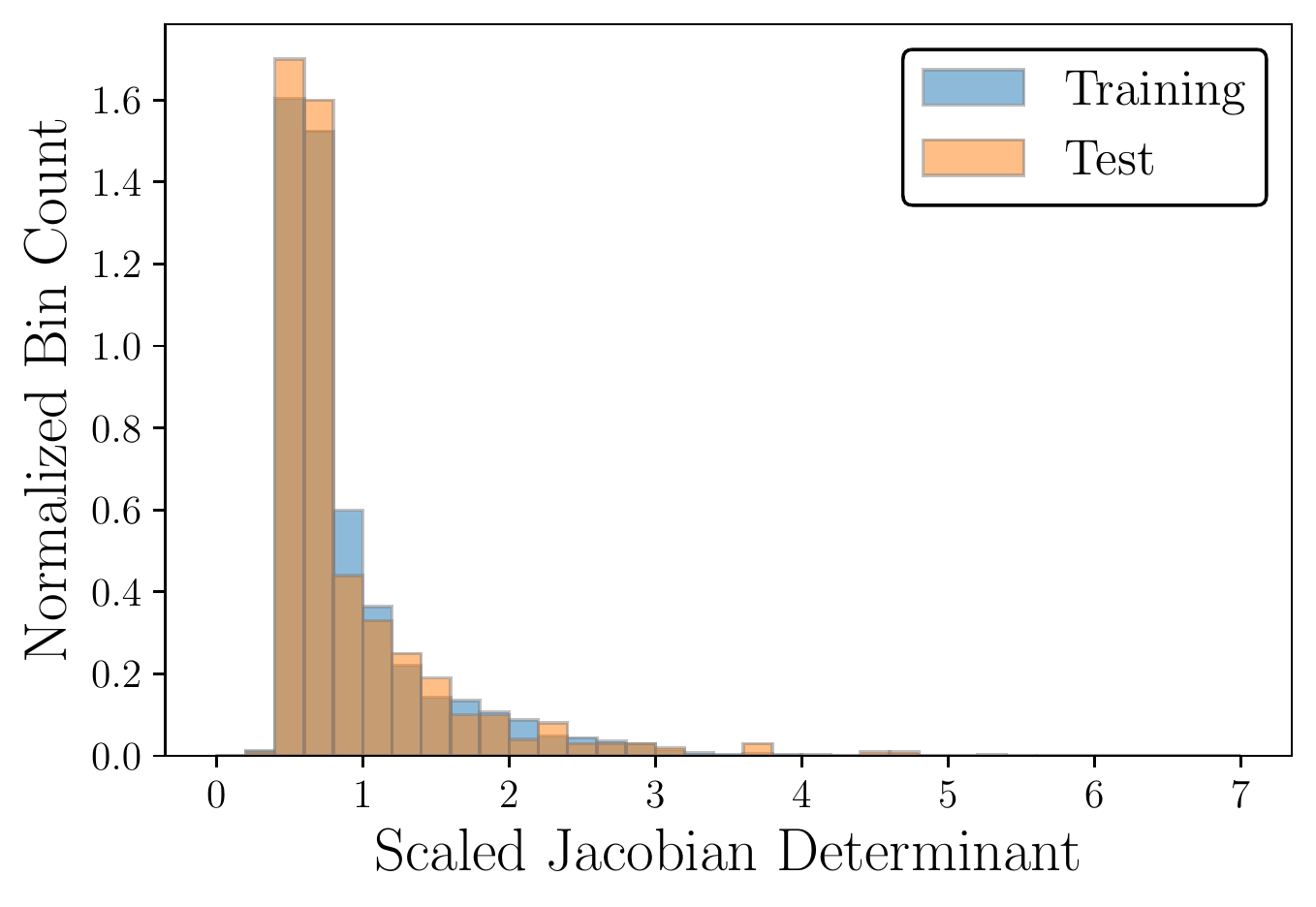}
			\caption{}\label{fig:scaled_jacobians_forward}
		\end{subfigure}
		\hfill
		\begin{subfigure}{0.48\textwidth}
			\centering
			\includegraphics[width=\linewidth]{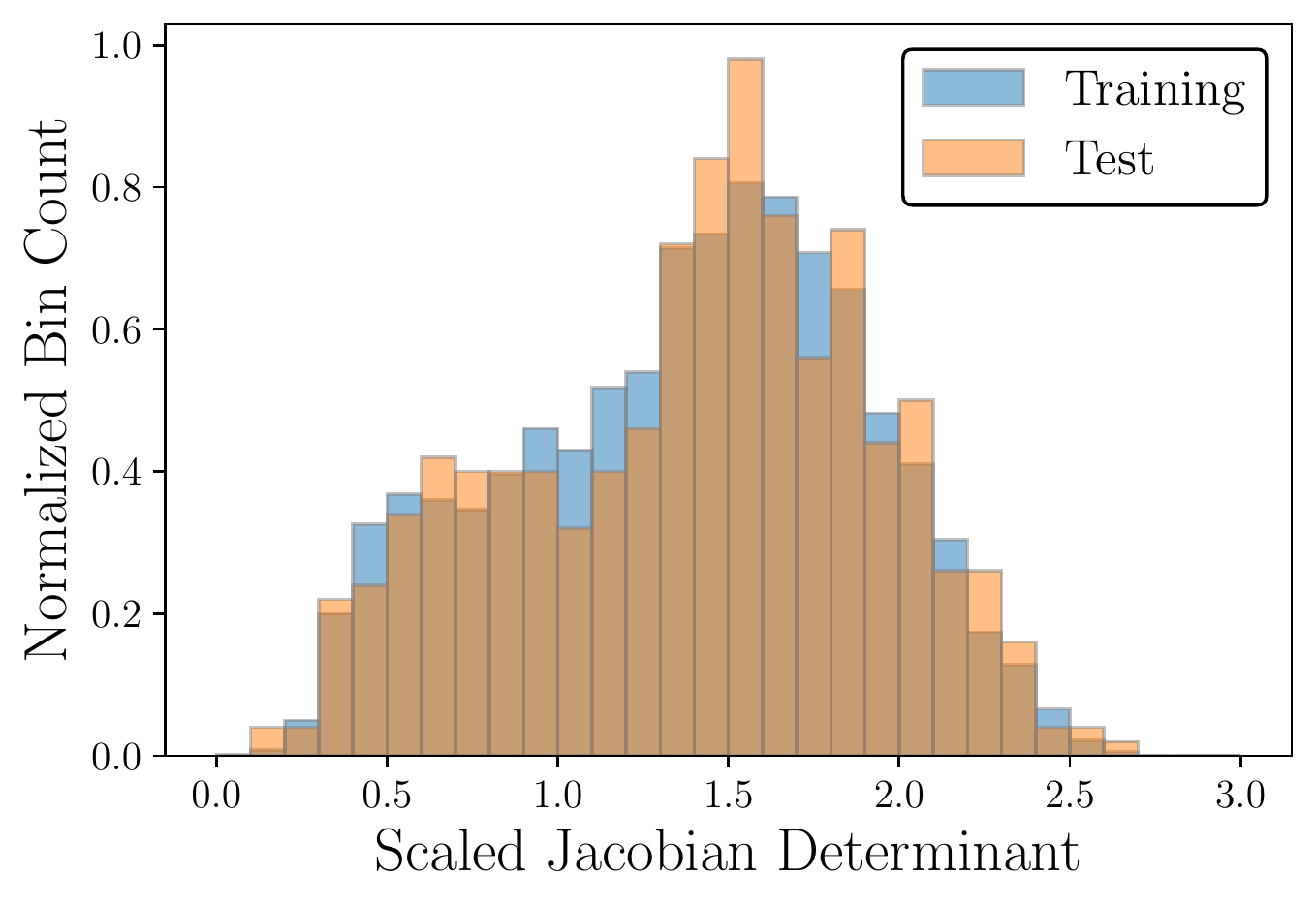}
			\caption{}\label{fig:scaled_jacobians_reverse}
		\end{subfigure}
		\caption{Histogram of (a) the quantity $J_h(\vect{\phi}_i)$ on our data-driven reduced coordinates and (b) the quantity $J_{h^{-1}}(\vect{\beta}_i)$ on the analytical reduced coordinates, after scaling to account for the order-of-magnitude difference. All bin counts have been normalized such that the plots correspond to empirical probability densities.}\label{fig:scaled_jacobians}
	\end{figure}

	We used both geometric harmonics (GHs) and neural networks (NNs) to fit $\vect{\beta}=h(\vect{\phi})$ and $\vect{\phi}=h^{-1}(\vect{\beta})$, achieving a high degree of predictive accuracy in both directions. The mean squared prediction errors are presented in Table~\ref{tab:rms_prediction_errors}. For both maps fit via GHs, we computed the Jacobian of the gradient at each input setting. Since values of the analytical effective parameters are approximately two orders of magnitude greater than those of the data-driven parameters, we scale the partial derivatives by the ratio of the corresponding coordinates' standard deviations over the data. This step removes the effects of scale differences between the two spaces. Figure~\ref{fig:scaled_jacobians} illustrates that the determinants are all of the same sign and are bounded away from zero and infinity.
	
	\begingroup
	\renewcommand*{\arraystretch}{1.33}
	\begin{table}[htb]
		\centering
		\begin{tabular}{c|rrr|rrr|}
			\cline{2-7}
			& \multicolumn{3}{c|}{$\vect{\beta}=h(\vect{\phi})$} & \multicolumn{3}{c|}{$\vect{\phi}=h^{-1}(\vect{\beta})$} \\
			& \multicolumn{1}{c}{$\beta_1$} & \multicolumn{1}{c}{$\beta_2$} & \multicolumn{1}{c|}{$\beta_3$} & \multicolumn{1}{c}{$\phi_1$} & \multicolumn{1}{c}{$\phi_5$} & \multicolumn{1}{c|}{$\phi_{11}$} \\ \hline
			\multicolumn{1}{|c|}{GH Train} & $1.10\times10^{-4}$ & $5.26\times10^{-4}$ & $4.02\times10^{-4}$ & $1.17\times10^{-6}$ & $1.61\times10^{-5}$ & $7.14\times10^{-5}$ \\
			\multicolumn{1}{|c|}{GH Test} & $1.87\times10^{-4}$ & $9.44\times10^{-4}$ & $7.04\times10^{-4}$ & $1.62\times10^{-6}$ & $2.36\times10^{-5}$ & $9.22\times10^{-5}$ \\
			\multicolumn{1}{|c|}{NN Train} & $1.34\times10^{-4}$ & $3.34\times10^{-4}$ & $2.98\times10^{-4}$ & $2.87\times10^{-5}$ & $9.86\times10^{-5}$ & $3.96\times10^{-4}$ \\
			\multicolumn{1}{|c|}{NN Test} & $1.47\times10^{-4}$ & $4.85\times10^{-4}$ & $3.99\times10^{-4}$ & $2.33\times10^{-5}$ & $9.85\times10^{-5}$ & $3.70\times10^{-4}$ \\ \hline
		\end{tabular}
		\caption{Root-mean-square prediction errors, by geometric harmonics and neural networks, of analytical and data-driven effective parameters for the compartmental model defined in Equation~(\ref{eqn:cole_model}).}
		\label{tab:rms_prediction_errors}
	\end{table}
	\endgroup
	
\begin{figure}[htb]
		\centering
		\includegraphics[width=\textwidth]{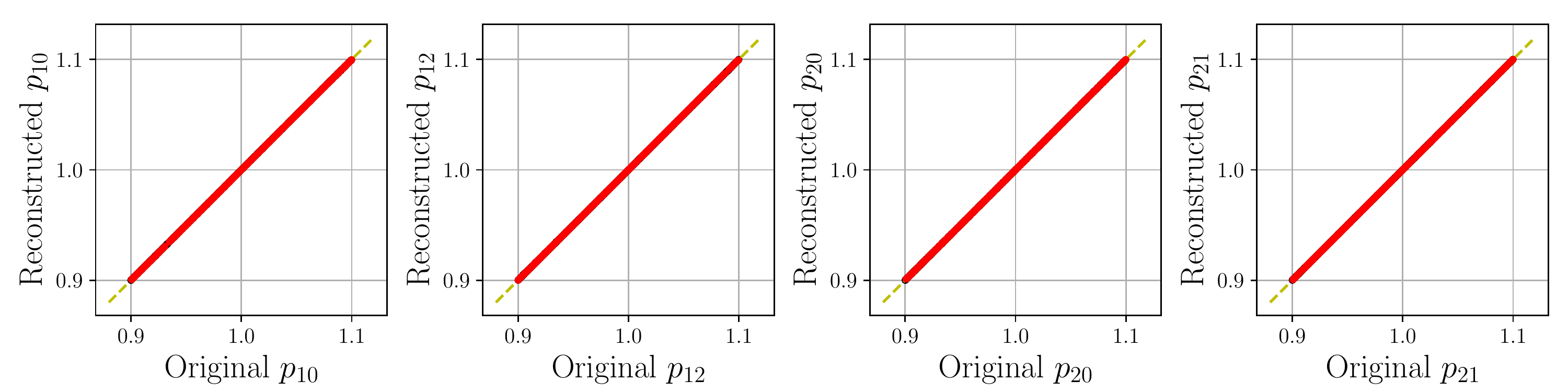}
		\caption{Comparison of the true parameter values $\vect{p}$ and the values recovered after applying the trained encoder and decoder, $f_d(f_e(\vect{p}))$. Test data, in red, overlay training data, in black.}
		\label{fig:cae_reconstruction.png}
	\end{figure}

	We also used a conformal autoencoder (CAE) to learn the effective and redundant parameter combinations. In particular, we trained subnetworks for the following three mappings:
	\begin{align}
		\text{Encoder:} & \qquad f_e:\R^4\to\R^4:\vect{p}\mapsto(\vect{\nu},\psi)\,,\\
		\text{Decoder:} & \qquad f_d:\R^4\to\R^4:(\vect{\nu},\psi)\mapsto\vect{p}\,,\\
		\text{Predictor:} & \qquad f_p:\R^3\to\R^{10}:\vect{\nu}\mapsto\vect{y}\,.
	\end{align}
	Figure~\ref{fig:cae_reconstruction.png} illustrates the trained CAE's ability to reconstruct the original parameter values with a high level of accuracy.
	
\begin{figure}[htbp]
		\centering
		\includegraphics[scale=1.0]{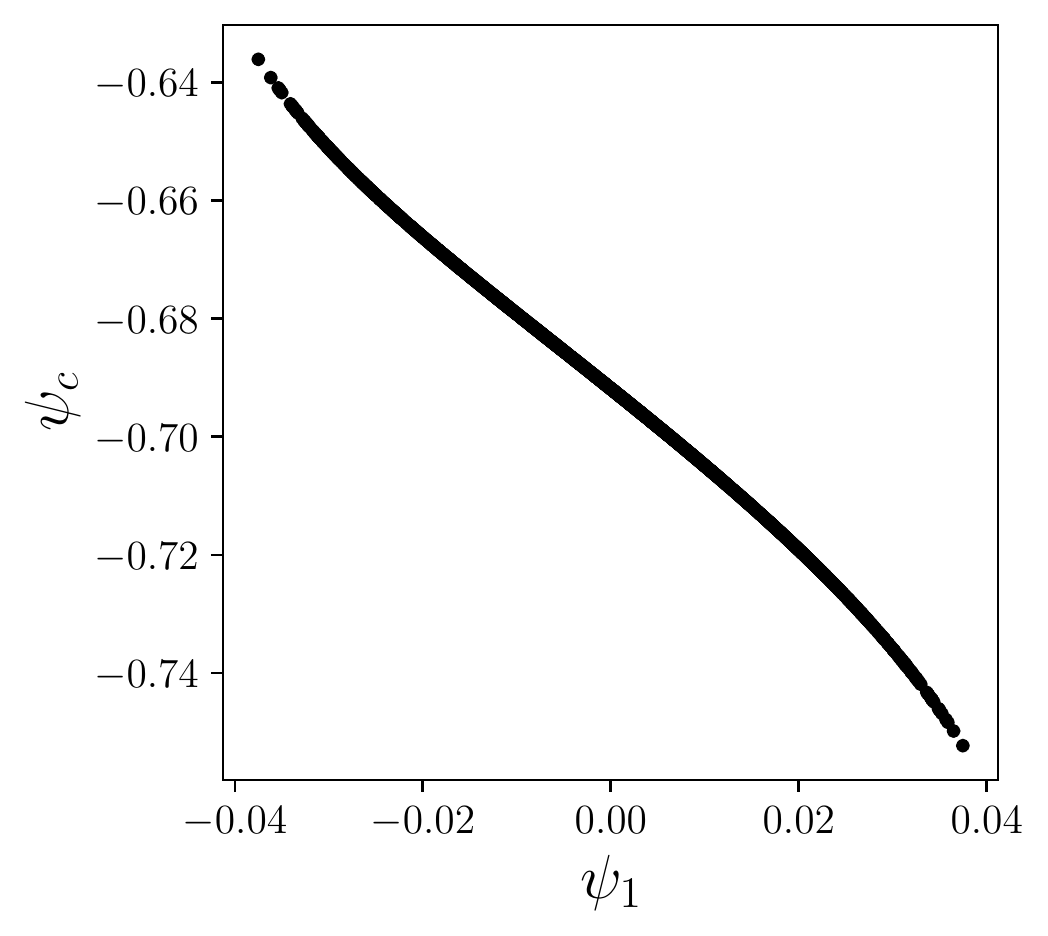}
		\caption{Comparison of two parameterizations for the redundant parameter of the compartmental model. The DMaps coordinate $\psi_1$ is obtained from 5000 new parameter vectors, $\vect{p}_i''$, chosen to achieve the same output behavior, $\tilde{\vect{y}}=\vect{y}(\tilde{\vect{p}})$, as our original base point. Our conformal autoencoder learns the quantity $\psi_c$ as the fourth output of its encoder subnetwork. Values of $\psi_c$ as predicted by the encoder for each $\vect{p}_i''$ are related to their corresponding DMaps coordinates by a one-to-one mapping.}
		\label{fig:vs_optimization_coordinate}
	\end{figure}

	We then sought to parameterize a level set of the redundant parameter $\psi$. We generate ${N''=5000}$ new parameter vectors, $\{\vect{p}_i''\}_{i=1}^{N''}$, as before; this time we perturb each component within $\pm25\%$ of its base value. From each of these points, we use the BFGS algorithm to minimize the objective
	\begin{equation}
		g(\vect{p})=\sum_{i=1}^{10}\bigg(y_i(\vect{p})-\tilde{y}_i\bigg)^2\,,\label{eqn:redundant_parameter_objective}
	\end{equation}
	with stopping criterion $\norm{\nabla g(\vect{p})}_\infty<10^{-8}$. The resulting minimizers of $g$ are vectors of parameters that achieve the same output as our base point $\tilde{\vect{p}}$. We compute DMaps on these minimizers and find that they lie on a one-dimensional manifold in the four-dimensional parameter space, corroborating our finding that there are three effective parameters, since $3+1=4$. We demonstrate in Figure~\ref{fig:vs_optimization_coordinate} that the DMaps coordinate $\psi_1$, which parameterizes the level set we discovered by optimizing $g$, is one-to-one with the redundant conformal coordinate of our CAE. In this case, we evaluated the trained encoder $f_e$ on each of the minimizers from our optimization data, which were not used during model training, and compared the fourth conformal coordinate against $\psi_1$.
	
	\section{Transitions Between Different Parametric Regimes: A Reaction Engineering Example}\label{sec:eta_phi_biot}
	Our last example is a static (steady state) problem from chemical kinetics, which allows us to discuss how our data-driven framework identifies the number of effective parameters when the system's response changes its nature (and dimensionality) in different parameter regimes. We consider a first-order catalytic reaction occurring in a spherical pellet with external mass transfer limitations. The parameters are  $k$ the reaction rate constant with units of inverse time $[1/\text{T}]$, $\alpha$ the characteristic length (pellet radius in the case of a sphere) with units of length $[\text{L}]$, $D$ the diffusion coefficient with units of length squared and per time $[\text{L}^2/\text{T}]$, and $k_m$ the external mass-transfer coefficient with units of length per time $[\text{L}/\text{T}]$. The output is the pellet production rate, expressed in the form of what is called the \emph{effectiveness factor}  $\eta$~\cite{rawlings2002chemical,thiele1939relation} defined as
	\begin{equation}
		\eta \equiv \frac{R_{jp}}{R_{jb}};
	\end{equation}
	Here, $R_{jp}$ is the pellet's production rate for the $j$-the species and $R_{jb}$ is the production rate for the the $j$-th species if the pellet reacted at the bulk concentration of this species both with units of moles per time per unit catalyst volume  $[\text{mol}/(\text{L}^3\cdot\text{T})]$.
	
	It can be shown~\cite{rawlings2002chemical} that
	\begin{equation}
		\eta =\frac{1}{\Phi}\bigg[\frac{1/\tanh(3\Phi)-1/(3\Phi)}{1+\Phi(1/\tanh(3\Phi)-1/(3\Phi))/\biot}\bigg]\,,
		\label{eq:effectiveness_factor}
	\end{equation}
	where $\Phi$ is the so-called Thiele modulus, and $\biot$ is the Biot number. For a first-order reaction, these dimensionless numbers are given by
	\begin{equation}
		\Phi\equiv\sqrt{\frac{k\alpha^2}{D}}\,,\qquad\biot \equiv \frac{k_m\alpha}{D}\,,
		\label{eq:thiele_and_biot}
	\end{equation}
	As can be seen from Equations~(\ref{eq:effectiveness_factor}) and~(\ref{eq:thiele_and_biot}), the response is effectively one-dimensional and depends on four parameters ($k,\alpha,D,k_m$). Since dimensional analysis already reduces the original number of parameters of the problem to only two---namely, $\Phi$ and $\biot$---this allow us to plot and visualize how the response changes for a range of these quantities. 
	
	For our computations we generated 10,000 pairs of $\Phi$ and $\biot$ independently and log-uniformly such that $\Phi \in [10^{-2}, 10^{6}]$ and $\biot \in {[10^{-4}, 10^{8}]}$ and computed the corresponding reaction rates and overall effectiveness factor values, Equation~(\ref{eq:effectiveness_factor}).
	\begin{figure}[htb]
		\centering
		\begin{subfigure}{0.48\textwidth}
			\centering
			\includegraphics[scale=0.55]{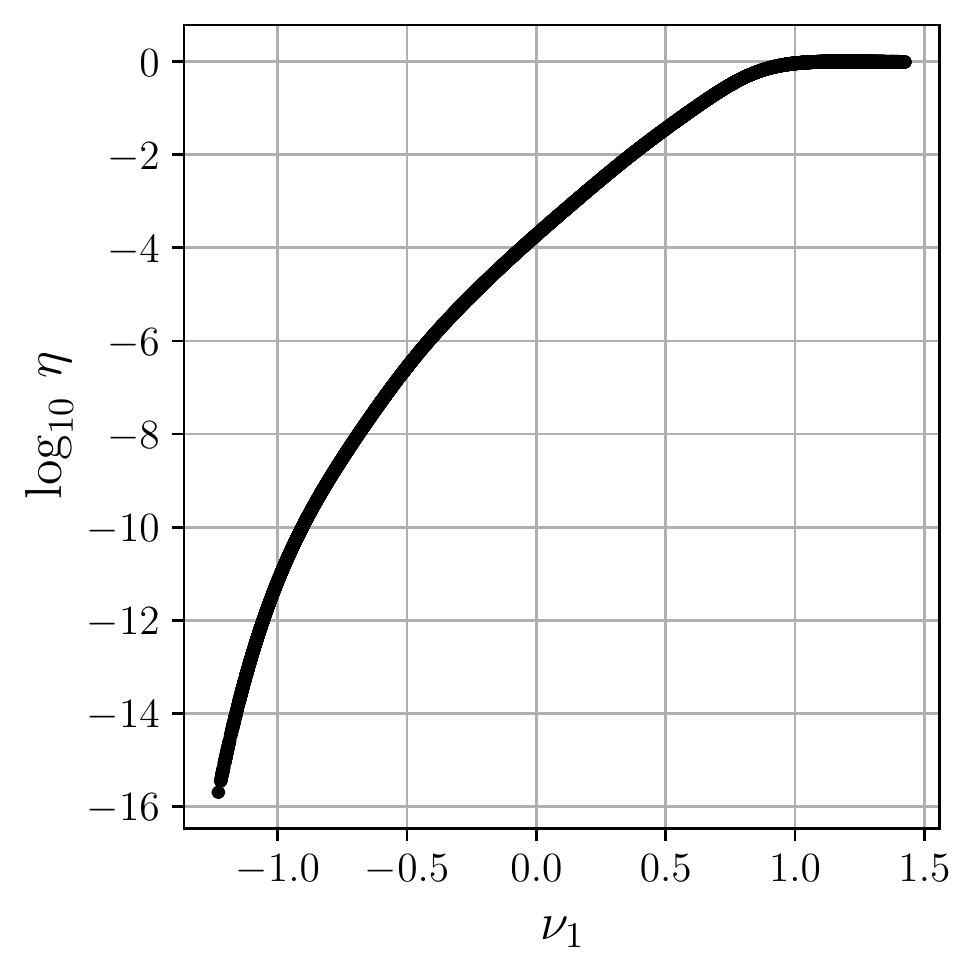}
			\caption{}\label{fig:effectiveness_factor_parity_plot}
		\end{subfigure}
		\hfill
		\begin{subfigure}{0.48\textwidth}
			\centering
			\includegraphics[scale=0.7]{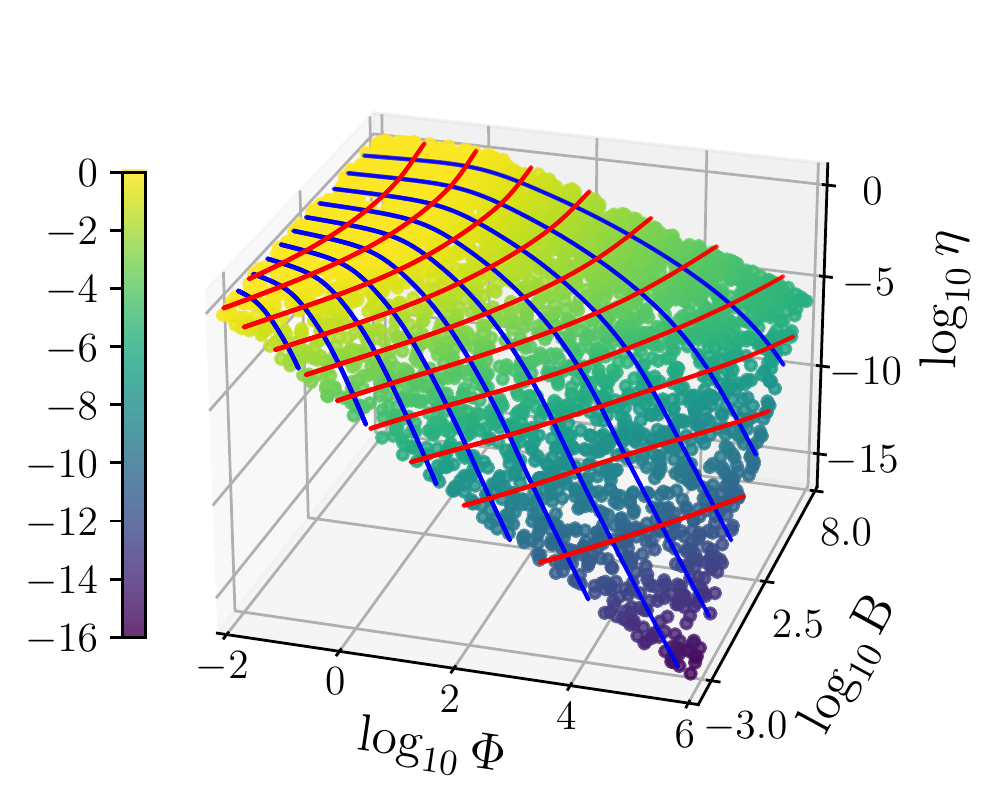}
			\caption{}\label{fig:effectiveness_factor_3d_plot}
		\end{subfigure}
		\caption{(a) The effective parameter $\nu_1$, as identified from the network, plotted against the effective factor computed from Equation~(\ref{eq:effectiveness_factor}). The true response surface of $\eta$ is plotted against the parameters $\Phi$ and $\biot$. The level sets of the meaningful parameter combination are shown with red lines and the redundant parameter combinations with blue lines. The values of $\eta$ for those level sets were predicted through the behavior estimator network.}\label{fig:effective_factor}
	\end{figure}
	We trained our Y-shaped conformal autoencoder network by using the same number of hidden layer and activation functions as the network discussed in Section~\ref{sec:Neural_Network}, but now only two input/bottleneck/output neurons. The effective parameter $\nu_1$ identified by the network, as can be seen in Figure~\ref{fig:figure1} on the left, is one-to-one with the analytical effective parameter $\eta$ (note also that it ``levels out'' in the region that the effectiveness factor becomes constant and equal to $1$, the so-called reaction-controlled regime). The JSF algorithm applied to this data also leads to a single data-driven effective parameter.
	
	Inspection of Equation~(\ref{eq:effectiveness_factor}) clearly shows that there are three qualitatively distinct parameter regimes depending on the magnitude of $\Phi$ and $\biot$: 
	\begin{equation}
		\eta(\Phi,\biot)\approx\left\{
		\begin{array}{cl}
			\biot/\Phi^2 & :\,\Phi>\max\left(\sqrt{\biot},\biot\right).\\
			1/\Phi & :\,1<\Phi<\biot\\
			1 & :\,\Phi<\min\left(\sqrt{\biot},1\right)\\
		\end{array}
		\right.
	\end{equation}
	We find that our framework is capable of capturing regime changes: in the first regime, where $\eta\approx\biot/\Phi^2$, the slopes of our level sets shown in Figure~\ref{fig:effectiveness_factor_3d_plot} show that both parameters affect the output, which therefore depends on a combination of both. In the second regime, where $\eta\approx1/\Phi$, our effective parameter level sets for $\eta$ become \emph{parallel to the $\biot$ number axis}, indicating that the effectiveness factor depends (locally) \emph{only} on $\Phi$. In the third (constant $\eta=1$) regime, our effective parameter correctly predicts the constant (the ``flattened out'' region we already pointed in in Figure~\ref{fig:effective_factor}). 
\end{document}